\documentclass[]{IEEEtran}
\usepackage{mathrsfs}
\usepackage{amsfonts}
\usepackage{graphicx,cite,epsfig,amssymb,amsmath}
\usepackage{color,xcolor}
\usepackage{pifont}
\usepackage{stmaryrd}
\usepackage{setspace}
\usepackage{subfigure}
\usepackage{cite}
\usepackage{array}
\usepackage{float}
\usepackage{multirow}
\usepackage[ruled,linesnumbered]{algorithm2e}
\usepackage{epstopdf}
\usepackage{mathtools}
\usepackage{multirow}
\usepackage{framed}
\usepackage{ntheorem}
\theoremseparator{.}
\theorembodyfont{\upshape}
\floatstyle{ruled}
\providecommand{\tabularnewline}{\\}
\providecommand{\algorithmname}{Algorithm}
\floatname{algorithm}{\protect\algorithmname}

\newtheorem{thm}{\bf Theorem}
\newtheorem{rem}{\bf Remark}
\newtheorem{lem}{\bf Lemma}

\newtheorem{coro}{\bf Corollary}
\newtheorem{defi}{\bf Definition}
\newtheorem{assu}{\bf Assumption}
\def\proof{{\emph{Proof:} }}

\begin{document}
\title{Accelerating DNN Training in Wireless Federated Edge Learning Systems}
\author{Jinke Ren, \emph{Graduate Student Member, IEEE}, Guanding Yu, \emph{Senior Member, IEEE}, and Guangyao Ding
 \thanks{Manuscript received Jul. 15, 2020; revised Sep. 19, 2020; accepted Oct. 23, 2020. \emph{(Corresponding author: Guanding Yu.)}}
	\thanks{J. Ren is with the College of Information Science and Electronic Engineering, Zhejiang University, Hangzhou 310027, China, and also with the Department of Electrical and Computer Engineering, Northwestern University, Evanston, IL 60208 USA (e-mail: renjinke@zju.edu.cn).}
    \thanks{G. Yu and G. Ding are with the College of Information Science and Electronic Engineering, Zhejiang University, Hangzhou 310027, China (e-mail: \{yuguanding, guangyaoding\}@zju.edu.cn).}}
\maketitle
\begin{abstract}
Training task in classical machine learning models, such as deep neural networks, is generally implemented at a remote cloud center for centralized learning, which is typically time-consuming and resource-hungry. It also incurs serious privacy issue and long communication latency since a large amount of data are transmitted to the centralized node. To overcome these shortcomings, we consider a newly-emerged framework, namely \textit{federated edge learning}, to aggregate local learning updates at the network edge in lieu of users' raw data. Aiming at accelerating the training process, we first define a novel performance evaluation criterion, called \textit{learning efficiency}. We then formulate a training acceleration optimization problem in the CPU scenario, where each user device is equipped with CPU. The closed-form expressions for joint batchsize selection and communication resource allocation are developed and some insightful results are highlighted. Further, we extend our learning framework to the GPU scenario. The optimal solution in this scenario is manifested to have the similar structure as that of the CPU scenario, recommending that our proposed algorithm is applicable in more general systems. Finally, extensive experiments validate the theoretical analysis and demonstrate that the proposed algorithm can reduce the training time and improve the learning accuracy simultaneously.
\end{abstract}

\section{Introduction}
With AlphaGo defeating the world's top Go player and the troika Y. Bengio, G. Hinton, and Y. LeCun winning the 2018 ACM A.M. Turing Award, \textit{artifical intelligence} (AI) becomes one of the most cutting-edge techniques in both academia and industry communities and is envisioned as a revolutionary innovation enabling a smart earth in the future \cite{Nature}. The implementation of AI in wireless networks is one of the most fundamental research directions, leading the trend of communication and computation convergence \cite{AI_in_Wireless_Survey_1,AI_in_Wireless_Survey_2,AI_in_Wireless_Survey_3}. The key idea of implementing AI in wireless networks is leveraging the rich data collected by massive distributed user devices to learn accurate AI models for network planning and optimization. Various conceptual and engineering AI breakthroughs have been applied to wireless network design, such as channel estimation \cite{Channel_Estimation}, signal detection \cite{Signal_Detection}, and resource allocation \cite{Resource_Allocation}.

In spite of the substantial progress in AI techniques, current learning algorithms demand enormous computation and memory resources for data processing. However, the training data in wireless networks is unevenly distributed over a large amount of resource-constrained user devices, whereas each device only owns a small fraction of data. Therefore, learning at devices suffers from isolated data island and will induce long training latency, making it hard to implement AI algorithms at user devices. Conventional solution generally offloads the local training data to a remote cloud center for centralized learning. Nevertheless, this method suffers from two key disadvantages. On the one hand, the latency for data transmission is often very large because of the limited communication resource. On the other hand, the privacy information involved in the training data may be leaked since the cloud center is inevitably attacked by some malicious third parties. Hence, traditional cloud-based learning framework is no longer suitable for the scenarios where data privacy is of paramount importance, such as intelligent healthcare systems and smart bank systems.

To address the first issue, an innovative architecture called \textit{mobile edge computing} (MEC) has been developed by implementing cloud computation capability at the network edge and migrating learning tasks from the cloud center to the edge server \cite{MEC_Survey}. By this means, the communication latency can be significantly reduced \cite{My_TWC}, the mobile energy consumption can be extensively saved \cite{You}, and the core network congestion can be notably relieved \cite{Mao}. To overcome the second deficiency, a novel distributed learning framework, namely \textit{federated learning} (FL) has been proposed. The key idea of this framework is to aggregate locally computed learning updates (gradients or parameters) in a centralized node while keeping the privacy-sensitive data remained at local devices \cite{Federated_Survey}. Toward this end, the benefit of the shared models trained from the rich data can be reaped and the computation resources of user devices can be exploited \cite{VR}. Motivated by this, effective collaboration between MEC and FL, referred to as \textit{federated edge learning} (FEEL) has the great potential to facilitate the implementation of AI in wireless networks \cite{Edge_Learning}.

As an important branch of FL, FEEL implements the centralized node at the wireless edge to relieve core network congestion. It requires the joint design of machine learning and wireless communication. This paper makes use of FEEL to accelerate the training task of general deep neural networks (DNN). We are inspired by the prior studies on applying FEEL to learn AI models\cite{Communication_Efficient_from_Decentralized_Data,Two_Strategies_for_FL,QSGD,Deep_Gradient_Compression,Gradient_Compression_Survey,Zhu,Non_IID,Multi_Task_Learning,Blockchain,Chen_1,BAA}. In \cite{Communication_Efficient_from_Decentralized_Data}, a federated averaging algorithm for distributed learning was developed, while both independent and identically distributed (IID) and non-IID datasets were used for performance evaluation. Two approaches called structured updates and sketched updates were proposed to reduce communication cost between edge server and user devices \cite{Two_Strategies_for_FL}. In particular, substantial studies devoted great effort to reducing the communication overhead by developing effective compression methods\cite{QSGD,Deep_Gradient_Compression,Gradient_Compression_Survey,Zhu}. Moreover, via creating a shared subset of data gathered from user devices, a collaborative learning strategy for non-IID data distributed system was developed to improve the learning accuracy \cite{Non_IID}. In addition, the authors of \cite{Multi_Task_Learning} investigated the multi-task system and proposed a system-aware optimization framework to balance the communication cost, stragglers, and fault tolerance. Besides, a block-chained FEEL architecture was introduced in \cite{Blockchain}, where the optimal block generation rate was derived to minimize the end-to-end latency. Last but not least, a broadband analog aggregation scheme based on over-the-air computation was proposed in \cite{BAA}, where two communication-and-learning trade-offs were developed to achieve a low-latency FEEL system.

The prior works on FEEL mainly focus on accelerating the training task from the communication perspective, i.e., reducing the communication overhead between centralized node and user devices. However, the major characteristics of wireless communications (e.g., channel fadings and resource scheduling) have not been considered, which may greatly affect the learning performance. A joint user selection and resource allocation policy towards minimizing the loss function was developed in \cite{Chen_1} by taking into account the impact of packet errors over wireless links. However, the optimization for the training task itself is also important and has not been investigated therein. To make progress, we consider the joint communication and computation resource allocation in this paper. In particular, both CPU and GPU scenarios are considered since DNN training is executed on either CPU or GPU. The major hyperparameter, i.e., the training batchsize is also optimized, which is of paramount importance to the learning performance improvement. This work is most related to the prominent FEEL works of \cite{Edge_Meets_Learning} and \cite{FL_over_Wireless_Networks}, where \cite{Edge_Meets_Learning} developed a control algorithm to achieve the trade-off between local update and global aggregation under fixed resource budget and \cite{FL_over_Wireless_Networks} proposed a joint communication and computation resource allocation policy to balance the training time and energy consumption. We go one further step towards the training batchsize optimization from the learning perspective. Our main result shows that the training batchsize should dynamically adapt to the wireless channel condition to achieve the most desirable learning performance. The main contributions of this work are summarized as follows.
\begin{itemize}
    \item To quantitatively analyze the training process, we first define a significant \textit{global loss decay} function with respect to (w.r.t) the training batchsize. A novel criterion, namely \textit{learning efficiency} is developed as the ratio of the global loss decay to the end-to-end latency, which can well evaluate the system learning performance.
    \item We consider the training acceleration problem in the CPU scenario under the communication and computation resource budgets. The closed-form solutions of joint batchsize selection and communication resource allocation are also derived. More precisely, the optimal batchsize increases linearly with the local training speed and increases sublinearly with the exponent of $-\frac{1}{2}$ with both the training priority ratio and the uplink data rate.
    \item We extend the training acceleration problem to the GPU scenario and develop a new function to characterize the relation between training latency and training batchsize for GPU modules. The corresponding solution in this scenario is proved to have the similar structure as that in the CPU scenario, revealing that our proposed algorithm can be applied to more general systems.
    \item We evaluate the performance of the proposed algorithms via extensive experiments using some popular DNN models and a real dataset. Numerical results demonstrate that our proposed scheme can achieve better learning performance than some benchmark schemes.
\end{itemize}

The rest of the paper is organized as follows. Section II introduces the FEEL system model, the DNN model, and the communication model. Section III analyzes the training process and formulates the training acceleration problem in the CPU scenario. The closed-form solution is developed in Section IV. Section V extends the training acceleration problem to the GPU scenario and discusses the solution in this case. Section VI presents the experimental results and the whole paper is concluded in Section VII.
\section{System Model}
\subsection{Federated Edge Learning System}
\begin{figure*}[htp]
\begin{center}
	\includegraphics[width=6.0in]{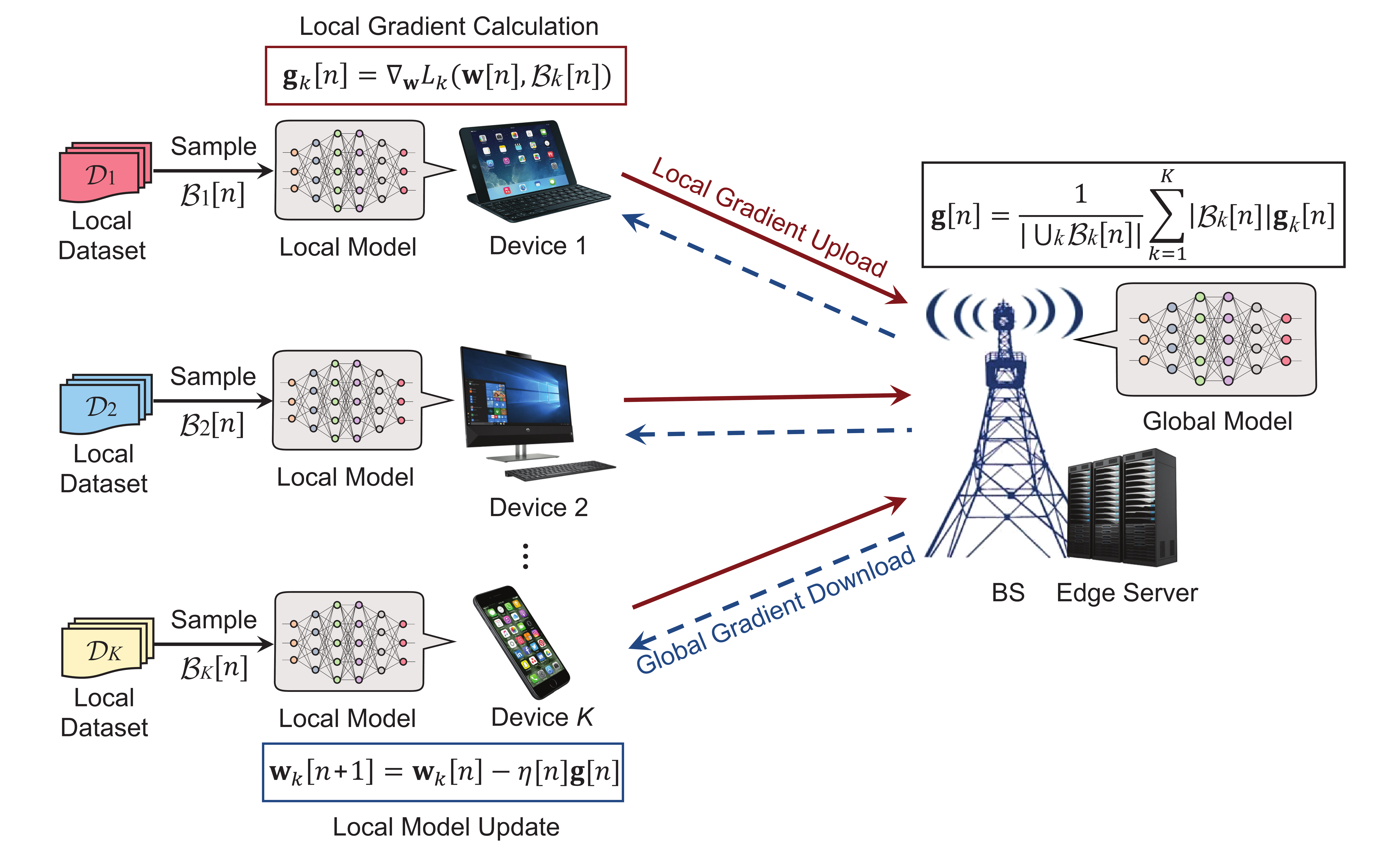}\\
	\caption{Federated edge learning system.}	
	\label{FEL system}
\end{center}
\end{figure*}

As depicted in Fig. \ref{FEL system}, we consider an FEEL system comprising one edge server and $K$ distributed single-antenna user devices, denoted by a set $\mathcal{K}=\{1, 2, \cdots, K\}$. Via interaction with its own user, each device collects a number of labeled data and constitutes its local dataset $\mathcal{D}_k=\left(({\bf{x}}_k^1, y_k^1),({\bf{x}}_k^2, y_k^2),\cdots, ({\bf{x}}_k^{N_k}, y_k^{N_k}) \right)$, where ${\bf{x}}_k^i$ is the $i$-th data sample of device $k$, $y_k^{i}$ represents the corresponding ground-truth label, and $N_k$ is the number of data that device $k$ owns. Considering the heterogeneous data structure, we assume that local datasets are statistically independent across user devices. A shared DNN model (e.g., convolutional neural network, CNN) is distributed over all devices, which needs to be collaboratively trained across the devices using their local datasets. To leverage the rich data as well as preserve data privacy, we adopt the federated learning technique by iteratively aggregating the locally computed gradients at the edge server.\footnote{We focus on the gradient-averaging implementation in the subsequent exposition as the gradients can be significantly compressed by some compression methods\cite{Deep_Gradient_Compression, Gradient_Compression_Survey}.} We introduce the learning mechanism by describing the steps that the communication parties take. The combination of the following steps is also referred to as a \textit{communication round}.
\begin{itemize}
    \item \textit{Step 1 (Local Gradient Calculation):}
     In each communication round, say the $n$-th round, each device selects a subset of data, denoted by $\mathcal{B}_k[n]$ from the local dataset $\mathcal{D}_k$, performs the backpropagation algorithm, and then computes the local gradient ${\bf{g}}_k[n]$.
	\item \textit{Step 2 (Local Gradient Upload):} After quantizing the local gradient, each device uploads the result to the edge server via multiple channel access, such as the time division multiple access (TDMA) or the orthogonal frequency division multiple access (OFDMA).
	\item \textit{Step 3 (Global Gradient Aggregation):} The edge server receives the local gradients from all devices and computes the global gradient as
           \begin{equation}
                {\bf{g}}[n] = \frac{1}{|\cup_k \mathcal{B}_{k}[n]|} \sum_{k=1}^{K} |\mathcal{B}_{k}[n]| {\bf{g}}_k[n]. \label{gradient aggregation}
            \end{equation}
    \item \textit{Step 4 (Global Gradient Download):} The edge server delivers the global gradient ${\bf{g}}[n]$ to the base station (BS), which transmits the global gradient to all devices.
    \item \textit{Step 5 (Local Model Update):}
         Each device updates the local model by
            \begin{equation}
                {\bf{w}}_k[n+1] = {\bf{w}}_k[n] - \eta[n] {\bf{g}}[n],~~\forall k, \label{local model update}
            \end{equation}
            where $\eta[n]>0$ is the learning rate in the $n$-th communication round.
\end{itemize}
Starting from $n = 1$, the edge server and user devices iterate the preceding steps until convergence.

\subsection{DNN Model}
In this work, we take a general DNN model for analysis. Let $p$ denote the total number of parameters in the DNN model. The local loss function of device $k$ that measures the training error on its dataset $\mathcal{D}_k$ is defined as
\begin{equation}
L_k\left({\bf{w}}\right)=\frac{1}{N_k} \sum_{(\scriptsize{{\bf{x}}},y) \in \mathcal{D}_{k}}\ell\left({\bf{w}},{\bf{x}},y\right),~~\forall k, \label{local loss function}
\end{equation}
where $\ell({\bf{w}},{\bf{x}},y)$ is the sample-wise loss function that quantifies the prediction error between the learning output (via data sample ${\bf{x}}$ and parameter set ${\bf{w}}$) and the ground-truth label $y$. Let $N=\sum_{k=1}^{K} N_k$. Then, the global loss function at the edge server can be expressed as
\begin{equation}
L\left({\bf{w}}\right)=\frac{1}{N} \sum_{k=1}^K  N_k L_k\left({\bf{w}}\right).\label{global loss fucntion}
\end{equation}
The main objective of the training task is to optimize the parameters towards minimizing the global loss function.

\subsection{Communication Model}
In this work, we adopt the TDMA method for data transmission.\footnote{Although we assume a TDMA scenario, the design principle and analytical framework are also applicable in other orthogonal and non-orthogonal multiple access methods.} Let $P_k^{\text{U}}$ denote the transmit power of device $k$ for gradient uploading and $h_k^{\text{U}}$ denote its uplink channel gain. Accordingly, define $P_k^{\text{D}}$ as the transmit power of the BS for transmitting the global gradient to device $k$ and $h_k^{\text{D}}$ as the downlink channel gain. To facilitate analysis, we assume that both channel gains are available at the BS in each communication round \cite{FL_over_Wireless_Networks}. 

The durations of uplink and downlink time-slots are relatively short (e.g., the frame duration of LTE protocol is 10 ms), during which the channel gains keep fixed. However, the duration of one communication round is often in the time-scale of second. It is because of the high computational complexity in running the backpropagation algorithm and the limited computation resource of user devices. Toward this end, each communication round will experience multiple time-slots. Since this work focuses on accelerating the training task from the long-term perspective, the channel dynamics would not affect the learning performance too much. Thus, we employ the average uplink and downlink data rates instead of the instantaneous ones, which are respectively evaluated as
\begin{align}
R_k^{\text{U}}= W \mathbb{E}_{h} \left\{\log_2 \left(1+\frac{P_k^{\text{U}} |h_k^{\text{U}}|^2}{N_0}\right)\right\},~~\forall k, \label{average upload data rate}\\
R_k^{\text{D}}= W \mathbb{E}_{h} \left\{\log_2 \left(1+\frac{P_k^{\text{D}} |h_k^{\text{D}}|^2}{N_0}\right)\right\},~~\forall k, \label{average download data rate}
\end{align}
where $\mathbb{E}_{h}$ represents the expectation over channel fading, $W$ denotes the system bandwidth, and $N_0$ implies the variance of complex white Gaussian channel noise  \cite{Wireless_Model,TVT}.

Thus far, we have elaborated the learning mechanism of FEEL and established both the learning and communication models. In the next section, we will analyze the training process and formulate the training acceleration problem. In our work, the edge server is always equipped with powerful GPU whereas the training module of user devices can be either CPU or GPU. Therefore, we will respectively investigate the CPU and GPU scenarios in the sequel.
\section{Federated Learning in CPU Scenario: Problem Formulation}
In this section, we consider the CPU scenario where each device is only equipped with CPU for training. We first analyze the training loss decay and the end-to-end latency in each communication round. Based on these, we formulate a training acceleration problem to maximize the system learning efficiency. 
\subsection{Training Loss Decay Analysis}
To accelerate the training process and maintain a desired learning accuracy, we take the mini-batch stochastic gradient descent (SGD) algorithm in this paper. The main challenge in performing this algorithm is the selection of the hyperparameter, i.e., the training batchsize, which greatly affects the learning accuracy and is valuable to be optimized.

To evaluate the training performance, we first define an auxiliary function, namely \textit{global loss decay}, as
\begin{equation}
\Delta L[n]= L \left({\bf{w}}[n]\right) - L \left({\bf{w}}[n+1]\right), \label{loss decay function}
\end{equation}
which represents the difference of the global loss function across the $n$-th communication round.

Now we analyze the training performance of the FEEL system. Since the five steps in one communication round will be iterated until the global model converges, we take one communication round for the following analysis. The index $n$ is omitted for brevity. In one communication round, each device selects a subset of data $\mathcal{B}_k$, namely one \textit{batch} to compute the local gradient, as ${\bf{g}}_k = \nabla_{\bf{w}} L_k \left({\bf{w}},\mathcal{B}_{k} \right)$. Then, the edge server aggregates all local gradients and computes the global gradient. Let $B_k=|\mathcal{B}_k|$ denote the number of data in device $k$'s batch. Thus, the FEEL system can process $B = \sum_{k=1}^{K} B_k$ data in one communication round, which is called \textit{global batchsize}. 
Then according to \cite{Loss_Decay_Batchsize}, the relation between the global loss decay function and the global batchsize can be approximately evaluated as
\begin{equation} \label{loss decrease function}
\Delta L= \xi \sqrt{B} = \xi \sqrt{\sum_{k=1}^{K} B_k},
\end{equation}
where $\xi$ is a coefficient determined by the specific DNN model. We note that the global loss decay does not increase linearly with the global batchsize. It is because the learning rate should dynamically adapt to the batchsize. In particular, we choose the scaling law of the learning rate $\eta=\mathcal{O}\left( \sqrt{B} \right)$. By this means, the variance in the gradient expectation keeps fixed to guarantee convergence and learning accuracy \cite{Learning_Rate_Adaptation1,Learning_Rate_Adaptation2}.
\subsection{End-to-End Latency Analysis}
As mentioned earlier, we aim at accelerating the training process. Therefore, the end-to-end latency is also an essential term to be optimized. In the following, we compute the dominant kinds of latency in one communication round.
\begin{itemize}
    \item \textit{Local Gradient Calculation Latency:} We measure the computation capability of each device by its CPU frequency, denoted by $f_k^{\text{C}}$ (in CPU cycles per second). Moreover, let $C$ denote the number of CPU cycles used for performing the backpropagation algorithm with one data sample. Since CPU operates in serial mode, the local gradient calculation latency can be expressed as
        \begin{equation}
            t_k^{\text{L,C}} = \frac{B_k C}{f_k^{\text{C}}},~~~\forall k.\label{Local Forward Propagation latency}
        \end{equation}
	\item \textit{Local Gradient Upload Latency:} The local gradient should be quantized before transmission. Since the computational complexity of gradient quantization is very low \cite{Gradient_Compression_Survey}, the quantization latency can be ignored as compared with the local gradient upload latency. Due to the fact that each parameter has a counterpart gradient, the total number of elements in each local gradient is equal to $p$. Let $b$ denote the average quantitative bit number for each gradient element. Then, the total data size of each local gradient is $s = b \times p$. Moreover, let $T_f^{\text{U}}$ denote the length of each uplink frame (usually 10 ms in LTE standards) and $\tau_k^{\text{U}}$ denote the time-slot duration allocated to device $k$ in one frame. Therefore, device $k$ can transmit $\tau_k^{\text{U}} R_k^{\text{U}}$ bits in one frame and the total number of frames that device $k$ needs to upload its gradient is $\frac{s}{\tau_k^{\text{U}} R_k^{\text{U}}}$.
Accordingly, the local gradient upload latency is given by
        \begin{equation}
            t_k^{\text{U}} = \frac{s T_f^{\text{U}}}{\tau_k^{\text{U}} R_k^{\text{U}}},~~\forall k. \label{Local Gradient Upload Latency}
        \end{equation}
    \item \textit{Global Gradient Download Latency:} Similar to the local gradient upload step, we use $b$ bits to quantize each global gradient element. To ensure that the end-to-end latency of different devices can be equalized and thus facilitate update synchronization, we also adopt TDMA for downlink channel access.\footnote{Our analytical framework can be extended to the broadcast model with a minor modification on the global gradient download latency, where the downlink data rate is determined by the minimum data rate among all devices.} Let $T_f^{\text{D}}$ denote the length of each downlink frame and $\tau_k^{\text{D}}$ denote the time-slot duration allocated to device $k$ in one frame. Then, the global gradient download latency is given by
        \begin{equation}
            t_k^{\text{D}} = \frac{s T_f^{\text{D}}}{\tau_k^{\text{D}} R_k^{\text{D}}},~~~\forall k. \label{Global Model Download Latency}
        \end{equation}
    \item \textit{Local Model Update Latency:} After receiving the global gradient, each device updates its local model according to (\ref{local model update}). Let $M^{\text{C}}$  denote the number of CPU cycles that are required for local model update. Then, the local model update latency can be expressed as
       \begin{equation}
            t_k^{\text{M,C}} = \frac{M^{\text{C}}}{f_k^{\text{C}}},~~\forall k. \label{Local Model Update Latency}
        \end{equation}
\end{itemize}

Assuming that the edge server is computationally powerful, we ignore the time for gradient aggregation. Moreover, each device uploads its gradient after it finishes local gradient calculation. Also, the edge server cannot start gradient averaging until it receives the local gradients from all devices. Toward this end, the end-to-end latency of device $k$ is given by
\begin{equation}
T_k = \max_{ k \in \mathcal{K}} \left\{t_k^{\text{L},\text{C}} + t_k^{\text{U}}\right\} + t_k^{\text{D}} + t_k^{\text{M,C}},~~\forall k.
\end{equation}
Accordingly, the end-to-end latency of one communication round is given by
\begin{align}
    T &= \max_{ k \in \mathcal{K}}T_k \\&= \max_{ k \in \mathcal{K}} \left\{t_k^{\text{L},\text{C}} + t_k^{\text{U}}\right\} + \max_{ k \in \mathcal{K}}  \left\{t_k^{\text{D}} + t_k^{\text{M,C}}\right\}. \label{CPU system latency}
\end{align}

\subsection{Problem Formulation}
In this paper, we aim at accelerating the training task, i.e., reducing the training time to minimize the global loss function $L\left({\bf{w}}\right)$. To better evaluate the training performance, we define a new criterion from the system perspective, as shown below.
\begin{defi} [\textnormal{Learning efficiency}]
In each communication round, the performance of the FEEL system can be evaluated by the \textit{learning efficiency}, which is defined as the ratio of the global loss decay to the end-to-end latency, i.e.,
\begin{equation}
E = \frac{\Delta L}{T}.
\end{equation}
\end{defi}
\begin{rem}
The learning efficiency reflects the global loss decrease rate in the training duration. We note that the objective of the training task is to seek an optimal model $\bf{w}^*$ that minimizes the global loss function $L\left(\bf{w}\right)$. Therefore, maximizing the learning efficiency is equivalent to minimizing the overall training time. Accordingly, learning efficiency is an appropriate metric to evaluate the training performance.
\end{rem}

Based on the preceding analysis, the objective of training acceleration can be transformed into learning efficiency maximization. Combining the results in (\ref{loss decrease function}) and (\ref{CPU system latency}), the optimization problem can be formulated as
\begin{subequations}
    \begin{eqnarray}
   \!\!\!\!\!\!\!\!\! \mathscr{P}_1: &\max\limits_{ \substack{ \{\tau_k^{\text{U}},\tau_k^{\text{D}},\\ B_k, B\}}} \!\! &\frac{\xi \sqrt{B}}{ \max \limits_{k \in \mathcal{K}} \left\{t_k^{\text{L,C}} + t_k^{\text{U}}\right\} +  \max\limits_{ k \in \mathcal{K}}  \left\{t_k^{\text{D}} + t_k^{\text{M,C}}\right\}}, \label{P1a}\\
    &{\text{s.t.}}&\sum_{k=1}^K \tau_k^{\text{U}} \le T_f^{\text{U}}, \label{P1b}\\
                    &&\sum_{k=1}^K \tau_k^{\text{D}} \le T_f^{\text{D}}, \label{P1c}\\
                    && \sum_{k=1}^K B_k = B, \label{P1d}\\
                    &&1 \leq B_k \leq B^{\text{max}},~\forall k \in \mathcal{K}, \label{P1e}\\
                    &&\tau_k^{\text{U}},~\tau_k^{\text{D}} \ge 0,~\forall k \in \mathcal{K}, \label{P1f}
    \end{eqnarray}
\end{subequations}
where (\ref{P1b}) and (\ref{P1c}) represent the uplink and downlink communication resource limitations, respectively, (\ref{P1d}) gives the total number of data that can be processed in one communication round, and (\ref{P1e}) bounds the minimum and maximum batchsizes, where $B^{\text{max}}$ is determined by the memory size and the CPU configuration of each device.
In particular, the durations of uplink and downlink frames are fixed according to LTE standards \cite{3GPP}. Moreover, to facilitate theoretical analysis, we relax $B_k$ into a continuous variable because $B^{\text{max}}$ is typically large, such as 128. Also, we can round up to the nearest integer for practical implementation.


\section{Federated Learning in CPU Scenario: Optimal Solution}
In this section, we first analyze the mathematical characteristics of problem $\mathscr{P}_1$ and then decompose it into two subproblems. The closed-form solutions for both subproblems are derived individually and some insightful results about network planning are also discussed.
\subsection{Problem Decomposition}
The main challenge in solving problem $\mathscr{P}_1$ is that the denominator of the objective function (\ref{P1a}) is non-smooth. To make progress, we first swap the numerator and denominator of (\ref{P1a}) and rewrite it as
\begin{equation}
    \min \limits_{\substack{\{\tau_k^{\text{U}},\tau_k^{\text{D}},\\ B_k, B\}}} \frac{\max \limits_{ k \in \mathcal{K}} \left\{ t_k^{\text{L,C}} + t_k^{\text{U}}\right\} + \max \limits_{ k \in \mathcal{K}} \left\{ t_k^{\text{D}} + t_k^{\text{M,C}} \right\}} {\xi \sqrt{B}}.
\end{equation}
It can be observed that the local gradient upload latency $t_k^{\text{U}}$ is determined only by the uplink time-slot $\tau_k^\text{U}$, while the global gradient download latency $t_k^\text{D}$ is determined only by the downlink time-slot $\tau_k^\text{D}$ and is independent of other variables. Meanwhile, $\tau_k^\text{U}$ and $\tau_k^\text{D}$ are generally independent of each other according to (\ref{P1b}) and (\ref{P1c}). Toward this end, one communication round can be divided into two subperiods. The first subperiod aims to perform local gradient calculation and uploading, which can be formulated as
\begin{equation}
    \begin{array}{l}
    \mathscr{P}_2:~\min\limits_{\{\tau_k^{\text{U}}, B_k, B\}} \max \limits_{k \in \mathcal{K}}   \left\{ \dfrac{ t_k^{\text{L,C}} + t_k^{\text{U}}} {\xi \sqrt{B}} \right\},\\
    ~~~~~~~~~~~~{\text{s.t.}}~~~~\text{(\ref{P1b}), (\ref{P1d}), (\ref{P1e}), and (\ref{P1f})}.
    \end{array} \label{P2}
\end{equation}
The second subperiod is to download global gradient and update local model, and thus can be formulated as
\begin{equation}
    \begin{array}{l}
    \mathscr{P}_3: ~\min\limits_{\{\tau_k^{\text{D}}, B\}}  \max \limits_{k \in \mathcal{K}}   \left\{ \dfrac{ t_k^{\text{D}} + t_k^{\text{M,C}}} {\xi \sqrt{B}} \right\},\\
    ~~~~~~~~~~{\text{s.t.}}~~~\text{(\ref{P1c}) and (\ref{P1f})}.
    \end{array}
\end{equation}
It needs to be emphasized that the value of $B$ in subproblem $\mathscr{P}_3$ should match that in subproblem $\mathscr{P}_2$. Therefore, the global batchsize is a global variable that can be optimized at last. 
\subsection{Solution to Subproblem $\mathscr{P}_2$}
We can observe that subproblem $\mathscr{P}_2$ is a min-max optimization problem and is hard to be solved directly. To make it better tractable, we first define $E^{\text{U}}$ as the maximum reciprocal of the uplink learning efficiency among all devices, i.e., $E^{\text{U}} = \max \limits_{k \in \mathcal{K}}  \left\{ \frac{ t_k^{\text{L,C}} + t_k^{\text{U}}}{\xi \sqrt{B}} \right\}$. Then, by parametric algorithm \cite{Parametric_Algorithm}, subproblem $\mathscr{P}_2$ can be transformed into
\begin{subequations}
    \begin{eqnarray}
   \mathscr{P}_4:&\min\limits_{\substack{\{\tau_k^{\text{U}}, B_k,\\ B, E^{\text{U}}\}}} &E^{\text{U}}, \label{P4a}\\
    &{\text{s.t.}} & t_k^{\text{L,C}} + t_k^{\text{U}} \leq\xi \sqrt{B} E^{\text{U}},~\forall k \in \mathcal{K}, \label{P4b}\\
    && \text{(\ref{P1b}), (\ref{P1d}), (\ref{P1e}), and (\ref{P1f})}. \label{P4c} \notag
    \end{eqnarray}
\end{subequations}
It can be easily verified that the constraint (\ref{P4b}) is non-convex, resulting in a non-convex optimization problem. 
To make progress, we first keep $B$ fixed and then determine the joint optimal batchsize selection and uplink resource allocation. When $B$ is fixed, problem $\mathscr{P}_4$ becomes a convex one, as presented in the following lemma.
\begin{lem} [\textnormal{Convexity}]
Given the value of $B$, problem $\mathscr{P}_4$ becomes a convex optimization problem.
\end{lem}

\proof
The proof is straightforward because when the value of $B$ is fixed, the objective function (\ref{P4a}), the constraints (\ref{P1b}), (\ref{P1d}), (\ref{P1e}), and (\ref{P1f}) are all convex. Moreover, (\ref{P4b}) is also convex since it is the summation of two convex functions. Therefore, $\mathscr{P}_4$ is convex for given value of $B$. The detailed derivation is omitted for brevity.

Lemma 1 is essential for solving problem $\mathscr{P}_4$ by applying the fractional optimization method. Moreover, classical convex optimization algorithms can be used. To better characterize the structure of the solution and gain more insightful results, we define two auxiliary indicators for each device, as
\begin{itemize}
    \item \textit{Local training speed} is defined as the speed of each device to perform backpropagation algorithm, i.e., $V_k=\frac{f_k^{\text{C}}}{C}$.
    \item \textit{Training priority ratio} is defined as the ratio of each local computation capability to the summation of all devices' computation capability, i.e., $\rho_k=\frac{f_k^{\text{C}}}{\sum_{k=1}^K f_k^{\text{C}}}$.
\end{itemize}
Then, the optimal solution to problem $\mathscr{P}_4$ with fixed $B$ can be described as follows.
\begin{thm}
The joint optimal batchsize selection and uplink resource allocation policy is given by
\begin{equation}
	\left\{
    \begin{aligned}
    &B_k^* =  \left[ \left( \Delta L E^{\text{U}*} - \left( \frac{\Delta L s T_f^{\text{U}} \mu^* }{\rho_k R_k^{\text{U}}}\right)^{\frac{1}{2}} \right) V_k \right]_{1}^{B^{\text{max}}}, \label{optimal batch} \\
    &\tau_k^{\text{U}*} = \left[ \frac{s}{R_k^{\text{U}}} \left( \Delta L E^{\text{U}*}-\frac{B_k^*}{V_k}  \right)^{-1} \right]^ + T_f^{\text{U}},
	\end{aligned}
    \right.~\forall k,
\end{equation}
\end{thm}
where $\mu^*$ and $E^{\text{U}*}$ are the optimal values satisfying the active time-sharing constraint $\sum_{k=1}^K \tau_k^{\text{U}*} = T_f^{\text{U}}$ and the global batchsize limitation $ \sum_{k=1}^K B_k^* = B$, respectively. Here, the operations $[X]_A^B = \min\left\{B, \max\left\{A, X\right\}\right\}$ and $[Y]^+ = \max \{Y, 0\}$.

\proof Please refer to Appendix A.
\begin{rem} [\textnormal{Threshold-based batchsize selection}]
Theorem 1 reveals that the optimal batchsize has a threshold-based structure and is mainly determined by three factors: the local training speed, the training priority ratio, and the uplink data rate. More precisely, the batchsize increases linearly with the local training speed and increases sublinearly with the exponent of $-\frac{1}{2}$ with both the training priority ratio and the uplink data rate. In particular, the batchsize of each device varies across different communication rounds because of the channel dynamics. This result is intuitive and plays an important role in hyperparameter tuning. On the one hand, it theoretically guides devices to increase the batchsize when the local training speed increases or the channel condition becomes better. On the other hand, the device with a higher training priority ratio is suggested to choose a larger batchsize since it is superior to speed up the model convergence.
\end{rem}
\begin{rem} [\textnormal{Adaptive resource allocation}]
Theorem 1 indicates that the optimal uplink resource allocation depends not only on the uplink data rate and the local training speed, but also on the training batchsize. Specifically, this solution guarantees that the edge server can receive all local gradients simultaneously. Therefore, the end-to-end latency of each device can be equalized and thus facilitate the update synchronization needed for gradient aggregation. This result also guides devices to balance the communication and computation costs, and thus can be regarded as a communication-and-computation trade-off. Concisely, the amount of time-slot resource decreases with the uplink data rate since the device with better channel quality requests less communication resource. In addition, when the local training speed decreases, the device should occupy more time-slot resources to reduce its end-to-end latency.
\end{rem}

We now determine the optimal values of $E^{\text{U}*}$ and $\mu^*$. Classical method is to perform the two-dimensional search algorithm. To facilitate the search process and reduce the computational complexity, we compute some useful bounds for these two parameters in the following. On the one hand, by considering a special case where each device uses the same batchsize as well as being allocated with identical time-slot resource, we can obtain an upper bound of $E^{\text{U}*}$. On the other hand, we relax the constraint (\ref{P1e}) and further apply the Karush-Kuhn-Tucker (KKT) conditions to achieve a lower bound of $E^{\text{U}*}$. With mathematical analysis, the range of $E^{\text{U}*}$ could be expressed in the following corollary. The detailed proof is provided in Appendix B.
\begin{coro}
The value of $E^{\text{U}*}$ follows
\begin{equation}
	\left\{
    \begin{aligned}
    &E^{\text{U}*} \geq \frac{1}{\Delta L} \left( \frac{B C}{\sum_{k=1}^K f_k^{\text{C}}} + s \left(\sum \limits_{k=1}^K \sqrt{\frac{\rho_k}{R_k^{\text{U}}}}\right)^2 \right),\\
    &E^{\text{U}*} \leq \max \limits_{k \in \mathcal{K}} \left\{ \frac{1}{\Delta L} \left(\frac{B}{K V_k} + \frac{K s}{R_k^{\text{U}}} \right) \right\}. \label{EU bound}
	\end{aligned}
    \right.
\end{equation}
\end{coro}

Based on Corollary 1, we further investigate another two special cases to determine the range of $\mu^*$. The first one corresponds to the scenario where the optimal batchsize for all devices are $B_k^* = 1,~\forall k$. The second one corresponds to the scenario of $B_k^*=B^{\text{max}},~\forall k$. Then the range of $\mu^*$ can be expressed in the following corollary. The detailed proof is presented in Appendix B.
\begin{coro}
When there exists at least one device whose optimal batchsize is in the interval $\left(1, B^{\text{max}} \right)$, the value of $\mu^*$ should satisfy
\begin{equation}
	\left\{
    \begin{aligned}
    &\mu^*  \geq \min \limits_{k \in \mathcal{K}} \left\{\dfrac{\left( \Delta L V_k E^{\text{U}*} - B^{\text{max}}\right)^2 \rho_k R_k^{\text{U}}}{\Delta L s T_f^{\text{U}} V_k^2}\right\}, \\
    &\mu^* \leq \max \limits_{k \in \mathcal{K}}  \left\{\dfrac{\left( \Delta L V_k E^{\text{U}*} - 1 \right)^2 \rho_k R_k^{\text{U}} }{\Delta L s T_f^{\text{U}} V_k^2}\right\}. \label{mu bound}
	\end{aligned}
    \right.
\end{equation}
\end{coro}
\begin{rem}
From Corollary 2, we can find that the values of $\mu^*$ and $E^{\text{U}*}$ are tightly coupled. Note that even this corollary is derived under the assumption that there is at least one device whose batchsize is between $1$ and $B^{\text{max}}$, it is still applicable since the special cases of $B_k^*=1$ and $B_k^*=B^{\text{max}},~\forall k$ rarely occur in practice. On the other hand, these two cases happen when $B=K$ and $B=K B^{\text{max}}$, where the solutions can be easily obtained via the KKT conditions.
\end{rem}

Based on the preceding discussion, we now develop an effective two-dimensional search algorithm to solve the subproblem $\mathscr{P}_2$ for a given value of $B$, as described in Algorithm 1. The main idea is to update the values of training batchsize and uplink time-slot in each iteration until the time-sharing constraint and the global batchsize limitation are both satisfied. Moreover, let $\epsilon$ denote the maximum tolerance for the one-dimensional search for both $\mu^*$ and $E^{\text{U}*}$. Then, the computational complexity of Algorithm 1 is $\mathcal{O}\left( \left(K\log\frac{1}{\epsilon}\right)^2\right)$, which can be easily implemented in practical systems.
\begin{algorithm}[t]
 \caption{Two-dimensional search algorithm for subproblem $\mathscr{P}_2$.}\label{algorithm}
 \KwIn{Uplink channel gain $h_k^{\text{U}}$, uplink frame $T_f^{\text{U}}$, and global batchsize $B$.}
 \KwOut{Time-slot allocation $\tau_k^{\text{U}*}$ and batchsize selection $B_k^*$.}

 Initialize the maximum tolerance $\epsilon$.\\
 Calculate the bounds $E_\ell^{\text{U}}$ and $E_h^{\text{U}}$ according to (\ref{EU bound}).\\
 Calculate $\sum\limits_{k=1}^K \tau_{k,\ell}^{\text{U}*}$ and $\sum\limits_{k=1}^K \tau_{k,h}^{\text{U}*}$ based on Theorem 1, involving the one-dimensional search for $\mu_\ell^*$ and $\mu_h^*$.\\
 \While{$\left|\sum\limits_{k=1}^K \tau_{k,\ell}^{\rm{U}*}-T_f^{\rm{U}} \right| > \epsilon$ $\rm{or}$ $\left|\sum\limits_{k=1}^K \tau_{k,h}^{\rm{U}*}- T_f^{\rm{U}} \right| > \epsilon$,}{
 Define  $ E_m^{\text{U}} = \dfrac{E_\ell^{\text{U}}+E_h^{\text{U}}}{2}$.\\
 Calculate $\sum_{k=1}^K \tau_{k,m}^{\text{U}*}$ with one-dimensional search for $\mu_m^*$.\\
 \eIf{ $\sum\limits_{k=1}^K \tau_{k,m}^{\rm{U}*} \geq T_f^{\rm{U}}$,}{$E_\ell^{\rm{U}} = E_m^{\rm{U}}$.}
 {$E_h^{\text{U}} = E_m^{\text{U}}$.}
 Update $\tau_{k,\ell}^{\text{U}*}$ and $\tau_{k,h}^{\text{U}*}$.
 }
\end{algorithm}
\subsection{Solution to Subproblem $\mathscr{P}_3$ and Global Discussion}
In this subsection, we solve the subproblem $\mathscr{P}_3$ and discuss the global solution to the original problem $\mathscr{P}_1$. Similarly, we define $E^{\text{D}}= \max \limits_{k \in \mathcal{K}}   \left\{\frac{ t_k^{\text{D}} + t_k^{\text{M,C}}}{\xi \sqrt{B}} \right\}$ as the maximum reciprocal of the downlink learning efficiency among all devices. Then, the subproblem $\mathscr{P}_3$ can be reformulated as
\begin{subequations}
    \begin{eqnarray}
   \mathscr{P}_5:\!\!\!&\min\limits_{\{\tau_k^{\text{D}}, B, E^{\text{D}}\}}&E^{\text{D}}, \label{P5a}\\
    &{\text{s.t.}} & t_k^{\text{D}} + t_k^{\text{M,C}} \leq \xi \sqrt{B}E^{\text{D}},~\forall k \in \mathcal{K}, \label{P5b}\\
    && \text{(\ref{P1c}) and (\ref{P1f})}. \label{P5c} \notag
    \end{eqnarray}
\end{subequations}
Given the value of $B$, the mathematical characteristic of problem $\mathscr{P}_5$ is similar to that of problem $\mathscr{P}_4$. Therefore, it can be solved using the KKT conditions. The closed-form solution is presented in the following theorem. The proof is similar to that of Theorem 1 and thus is omitted due to page limits.
\begin{thm}
The optimal downlink resource allocation policy is given by
\begin{equation}
    \tau_k^{\text{D}*} = \left[ \frac{s}{R_k^{\text{D}}} \left( \Delta L E^{\text{D}*}-\frac{M^{\text{C}}}{f_k^{\text{C}}}  \right)^{-1} \right]^ + T_f^{\text{D}},
\end{equation}
where $E^{\text{D}*}$ is the optimal value associated with the time-sharing constraint $\sum_{k=1}^K \tau_k^{\text{D}*} = T_f^{\text{D}}$.
\end{thm}
\begin{rem}[\textnormal{Consistent resource allocation}] The result in Theorem 2 is rather intuitive that more time-slot resources should be allocated to the device with worse downlink channel condition to achieve a smaller end-to-end latency. By this means, the local model update at each device can be accomplished simultaneously and thus lead to a synchronous FEEL system.
\end{rem}

As yet, we have obtained the closed-form expressions for joint batchsize selection and uplink/downlink time-slot allocation as the function of the global batchsize $B$. Applying these results makes problem $\mathscr{P}_1$ become an univariate optimization problem with only variable $B$, which can be effectively solved by classical gradient descent algorithm or one-dimensional bisection algorithm. Then, the overall computational complexity for solving $\mathscr{P}_1$ is $\mathcal{O}\left( \frac{1}{\sqrt{\epsilon}}\left(K\log\frac{1}{\epsilon}\right)^2\right)$.
\section{Extension to GPU Scenario}
In this section, we consider the scenario where each device is equipped with GPU for training. We first propose a new GPU training function and then extend the training acceleration problem in this case. In particular, the optimal solution is proved to have a similar structure as that in the CPU scenario.
\subsection{GPU Training Function}
\begin{figure}
	\centering
	\subfigure[Theoretical result.]{
		\includegraphics[width=3in]{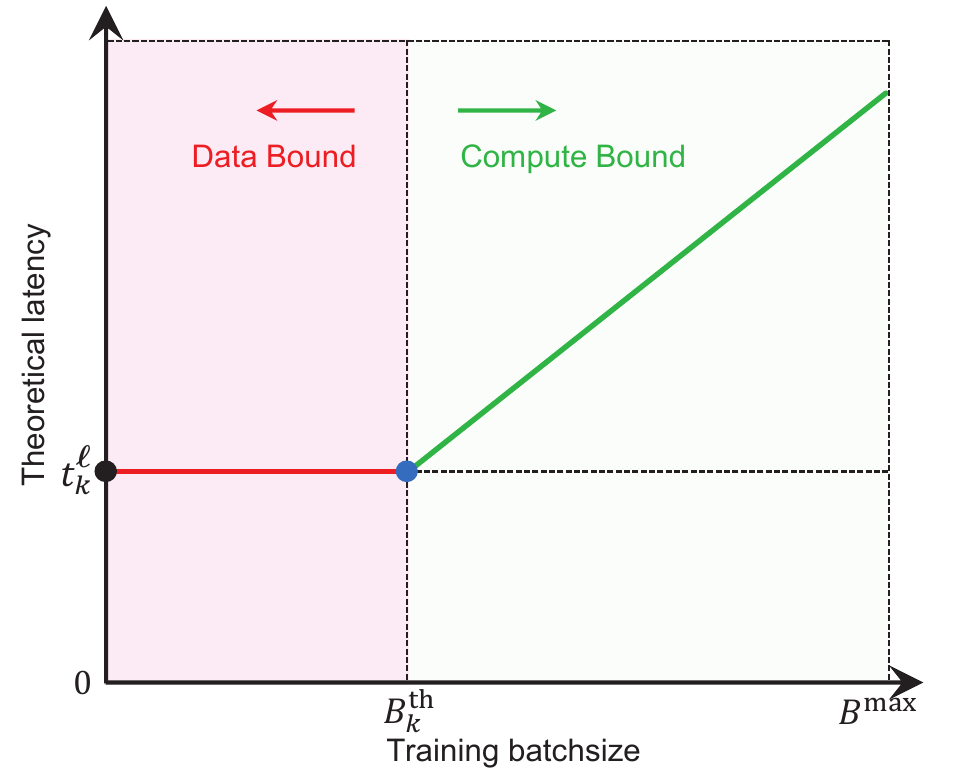}\label{GPU_theoretical}

	}
	\subfigure[Experimental result.]{
		\includegraphics[width=3in]{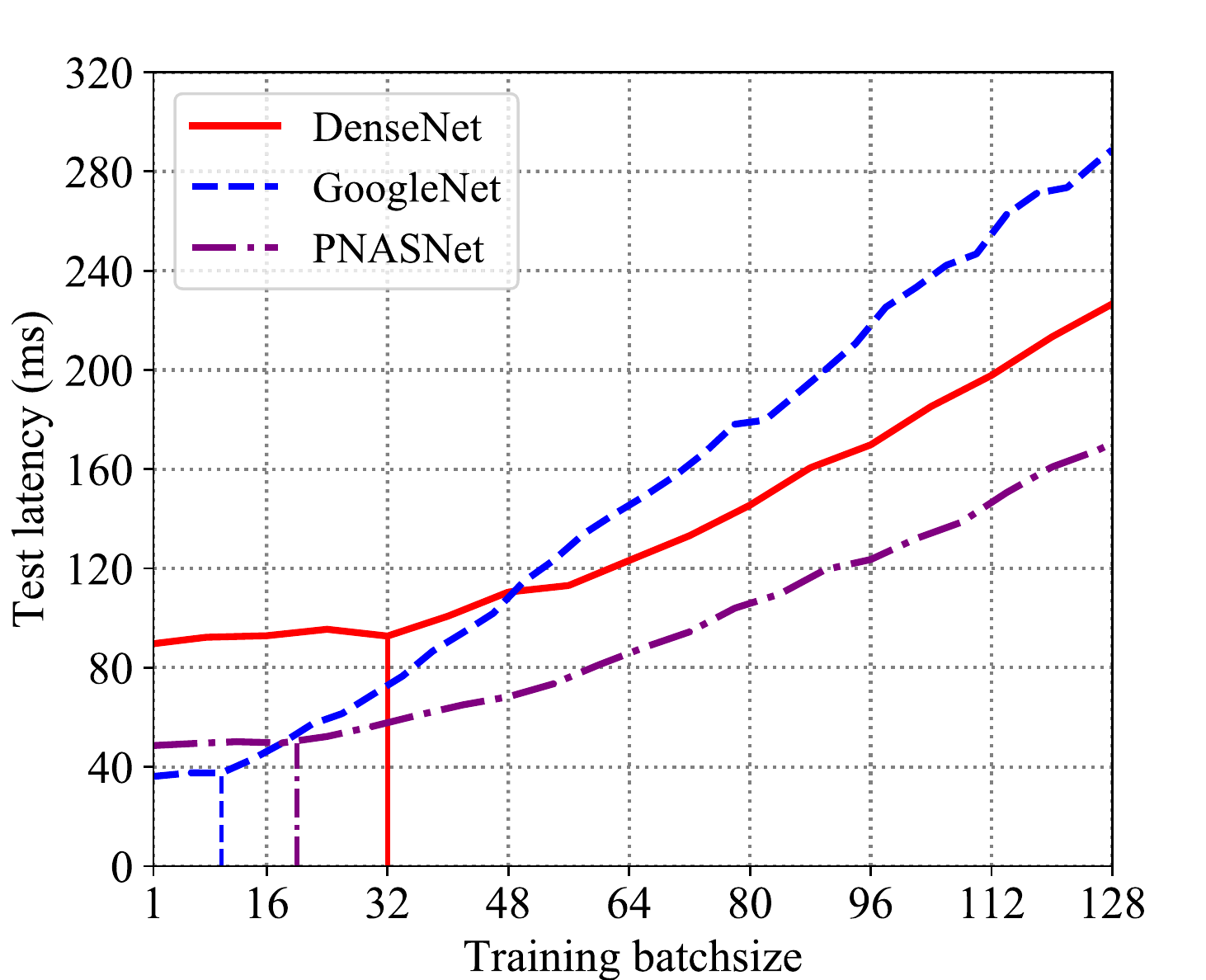}\label{GPU_experimental}
	}
	\caption{Local gradient calculation latency w.r.t. training batchsize.}
\end{figure}

Different from the serial mode of CPU, GPU execute in the parallel mode. Therefore, the local gradient calculation latency is no longer proportional to the training batchsize. Specifically, when the training batchsize is small, GPU can directly process all data simultaneously, leading to a constant training latency. In contrary, it grows once the training batchsize exceeds the maximum number of data that it can process at a time. With this consideration, we propose a new function to capture the relation between the local gradient calculation latency and the training batchsize, as presented in the following assumption and shown in Fig. \ref{GPU_theoretical}.
\begin{assu} [\textnormal{GPU training function}]
In the GPU scenario, the relation between the local gradient calculation latency and the training batchsize is given by
\begin{equation}
	t_k^{\text{L}, \text{G}} = \left\{
    \begin{aligned}
    &t_k^\ell,~~~~~~\text{if}~~1 \leq B_k \leq B_k^{\text{th}},\\
    &c_k \left(B_k - B_k^{\text{th}}\right) + t_k^\ell \triangleq t_k^h(B_k),   \\ &~~~~~~~~~~\text{if}~~B_k^{\text{th}}< B_k \leq B^{\text{max}},
	\end{aligned}
    \right. ~~~\forall k,\label{GPU Local Forward Propagation latency}
\end{equation}
where $t_k^\ell$, $c_k$, and $B_k^{\text{th}}$ are three coefficients determined by the specific DNN model (e.g., the number of layers and the number of neurons) and the GPU configuration of each device (e.g., the video memory size and the number of cores).
\end{assu}
\begin{rem}
Assumption 1 reveals that when the training batchsize is below a threshold $B_k^{\text{th}}$, the local gradient calculation latency keeps fixed because of GPU's parallel execution capability. In this case, the data samples are inadequate such that the computation resource (e.g., the video memory) is not fully exploited. Therefore, we name it as the \textit{data bound} region. On the other hand, the local gradient calculation latency grows linearly when the training batchsize exceeds the threshold. It is because in this case, the computation resource cannot support processing all data samples simultaneously. We note that the local gradient calculation latency does not increase in a ladder form. The reason is rather intuitive that the operations of data reading and transferring between memory modules and processing modules need additional time, leading to a comprehensive linear trend. Consequently, we name this region as the \textit{compute bound} region.
\end{rem}

To validate Assumption 1, we implement three popular DNN models: DenseNet, GoogleNet, and PNASNet on a Linux server equipped with three NVIDIA GeForce GTX 1080 Ti GPUs. The experimental results are depicted in Fig. \ref{GPU_experimental}. It can be observed that the local gradient calculation latency of each model first keeps invariant and then increases almost linearly with the training batchsize. This result shows that the curves obtained via experiments fit the theoretical model in (\ref{GPU Local Forward Propagation latency}) very well, which demonstrates the applicability of the proposed GPU training function.
\subsection{Problem Formulation and Analysis}
The learning efficiency in the GPU scenario is still defined as the ratio of the global loss decay to the end-to-end latency in one communication round.
In particular, the data sizes of both local gradient and global gradient in the GPU scenario are identical to those in the CPU scenario. Thus, the local gradient upload latency and global gradient download latency can be still expressed as (\ref{Local Gradient Upload Latency}) and (\ref{Global Model Download Latency}), respectively. Besides, let $f_k^{\text{G}}$ (in floating-point operations per second) denote the computation capability of device $k$ and define $M^{\text{G}}$ as the number of floating-point operations that are required for model update. Then, the local model update latency in this case is given by
\begin{equation}
            t_k^{\text{M,G}} = \frac{M^{\text{G}}}{f_k^{\text{G}}},~~\forall k. \label{Global Model Training Latency}
\end{equation}
Accordingly, the end-to-end latency of one communication round is given by
\begin{equation}
T = \max_{ k \in \mathcal{K}} \left\{t_k^{\text{L,G}} + t_k^{\text{U}} \right\} + \max_{k \in \mathcal{K}} \left\{t_k^{\text{D}} + t_k^{\text{M,G}}\right\}. \label{GPU system latency}
\end{equation}
Consequently, the training acceleration problem can be formulated as
\begin{equation}
    \begin{array}{l}
    \mathscr{P}_6:\max\limits_{\substack{\{\tau_k^{\text{U}}, \tau_k^{\text{D}},\\ B_k, B\}}}~  \dfrac{\xi \sqrt{B}}{\max \limits_{ k \in \mathcal{K}} \left\{ t_k^{\text{L,G}} + t_k^{\text{U}}\right\} + \max\limits_{k \in \mathcal{K}}  \left\{t_k^{\text{D}} + t_k^{\text{M,G}}\right\}}, \label{P6a}\\
    ~~~~~~~~~{\text{s.t.}}~~~\text{(\ref{P1b})--(\ref{P1f})}.
    \end{array}
\end{equation}

It is not easy to solve problem $\mathscr{P}_6$ since the local gradient calculation latency $t_k^{\text{L,G}}$ is non-differential. Traditional solution is the sub-gradient algorithm, which performs iterative optimization of the three-variable set, $\{\tau_k^{\text{U}},\tau_k^{\text{D}}, B_k\}$, via the sub-gradient functions. However, the computational complexity is too high. To address this issue, we analyze the mathematical characteristic of the objective function in (\ref{P6a}) and derive a necessary condition for the solution to problem $\mathscr{P}_6$, as described in the following lemma.
\begin{lem}
In the GPU scenario, the optimal batchsize $B_k^*$ of each device shall locate in the compute bound region, i.e., $B_k^{\text{th}} \leq B_k^* \leq B^{\text{max}}$.\footnote{In practical systems, the batchsize threshold $B_k^{\text{th}}$  is generally small. Therefore, we assume that each local dataset size is larger than the corresponding batchsize threshold.}
\end{lem}

\proof
Please refer to Appendix C.

The result in Lemma 2 coincides with the empirical result that the computation resource of all devices should be fully exploited to achieve the largest learning efficiency. Accordingly, the data bound region can be neglected and $\mathscr{P}_6$ can be reformulated as
\begin{subequations}
    \begin{eqnarray}
   \!\!\!\!\!\!\!\!\mathscr{P}_7:&\max\limits_{\substack{\{\tau_k^{\text{U}},\tau_k^{\text{D}},\\ B_k, B\}}} &\frac{\xi \sqrt{B}}{\max\limits_{ k \in \mathcal{K}} \left\{ t_k^h + t_k^{\text{U}}\right\} + \max \limits_{ k \in \mathcal{K}}  \left\{t_k^{\text{D}} + t_k^{\text{M,G}}\right\}},\\ \label{P7a}
   &{\text{s.t.}}  & B_k^{\text{th}} \leq B_k \leq B^{\text{max}},~~\forall k \in \mathcal{K},\label{P7c}\\
   && \text{(\ref{P1b})}, \text{(\ref{P1c})}, \text{(\ref{P1d})}, \text{and}~\text{(\ref{P1f})}.\notag
    \end{eqnarray}
\end{subequations}
With comprehensive comparison, we can observe that the structures of problems $\mathscr{P}_1$ and $\mathscr{P}_7$ are identical except the expressions of local gradient calculation latency. Fortunately, the two kinds of latency are proportional to the batchsize, making both problems essentially the same structure. Therefore, the algorithms for solving problem $\mathscr{P}_1$ are still applicable in solving problem $\mathscr{P}_7$. As a result, the solution in the GPU scenario has the similar structure as that in the CPU scenario. The detailed derivations are omitted due to page limits.

\section{Experiments}
\subsection{Experiment Settings}
In this section, we conduct experiments to validate our theoretical analysis and evaluate the performance of the proposed algorithms. The system setup is summarized as follows unless otherwise specified. We consider a single-cell network with a radius of $200$ m. The BS is located in the center of the network and each device is uniformly distributed in the cell. The channel gains of both uplink and downlink channels are generated following the pass-loss model: $PL~\text{[dB]} = 128.1 + 37.6 \log d$ [km], where the small-scale fading is set as Rayleigh distributed with uniform variance. The transmit powers of the uplink and downlink channels are both $28$ dBm. The system bandwidth $W = 10$ MHz and the channel noise density $N_0 = -174$ dBm/Hz. The lengths of each uplink and downlink frame are set as $T_f^{\text{U}} = T_f^{\text{D}} = 10$ ms according to the LTE standards. The average quantitative bit number for each gradient element is $b = 32$ bits and the maximum batchsize is bounded by $B^{\text{max}} = 128$ data samples. The detailed simulation parameters are listed in Table I.
\begin{table}[htp!]
\small
\caption{System Parameters}\label{parameter}
\centering{}
\vspace{-0.5em}
\begin{tabular}{|c|c|}
\hline
Parameter & Value\tabularnewline
\hline
\hline
Cell radius, $r$ & $200$~m \tabularnewline
\hline
System bandwidth, $W$ & $10$~MHz \tabularnewline
\hline
Noise power density, $N_0$ & $-174$~dBm/Hz \tabularnewline
\hline
Pass loss model &$128.1+37.6\log_{10}(d[\text{km}])$\tabularnewline
\hline
Time frames, $T_f^{\text{U}}, T_f^{\text{D}}$ & $10$~ms \tabularnewline
\hline
Transmit power, $P_k^{\text{U}}, P_k^{\text{D}}$& $28$~dBm \tabularnewline
\hline
Quantitative bit number, $b$, &$32$ bits\tabularnewline
\hline
Maximum batchsize, $B^{\text{max}}$ &  $128$  \tabularnewline
\hline
\end{tabular}
\end{table}

For exposition, we choose three popular DNN models: DenseNet121, ResNet18, and MobileNetV2 for image classification, where their number of parameters are 8,062,504, 33,161,024, and 3,538,984, respectively. The well-known dataset CIFAR-10 is used for model training, which consists of 50,000 training images and 10,000 validation images of 32 $\times$ 32 pixels in 10 categories.
Moreover, to simulate the distributed mobile data, we study two typical ways of partitioning data over devices \cite{BAA}: 1) IID case, where all data samples are first shuffled and then partitioned into $K$ equal parts, and each device is assigned with one particular part; 2) non-IID case, where all data samples are first sorted by digit label and then divided into $2K$ shards of size $25000/K$, and each device is assigned with two shards. Note that the latter is a pathological non-IID partition way since most devices obtain the data samples with only two types of digits.
\subsection{CPU Scenario: Generalization Ability with Different DNN Models}
\begin{figure}
	\centering
	\subfigure[Global training loss]{
		\includegraphics[width=3in]{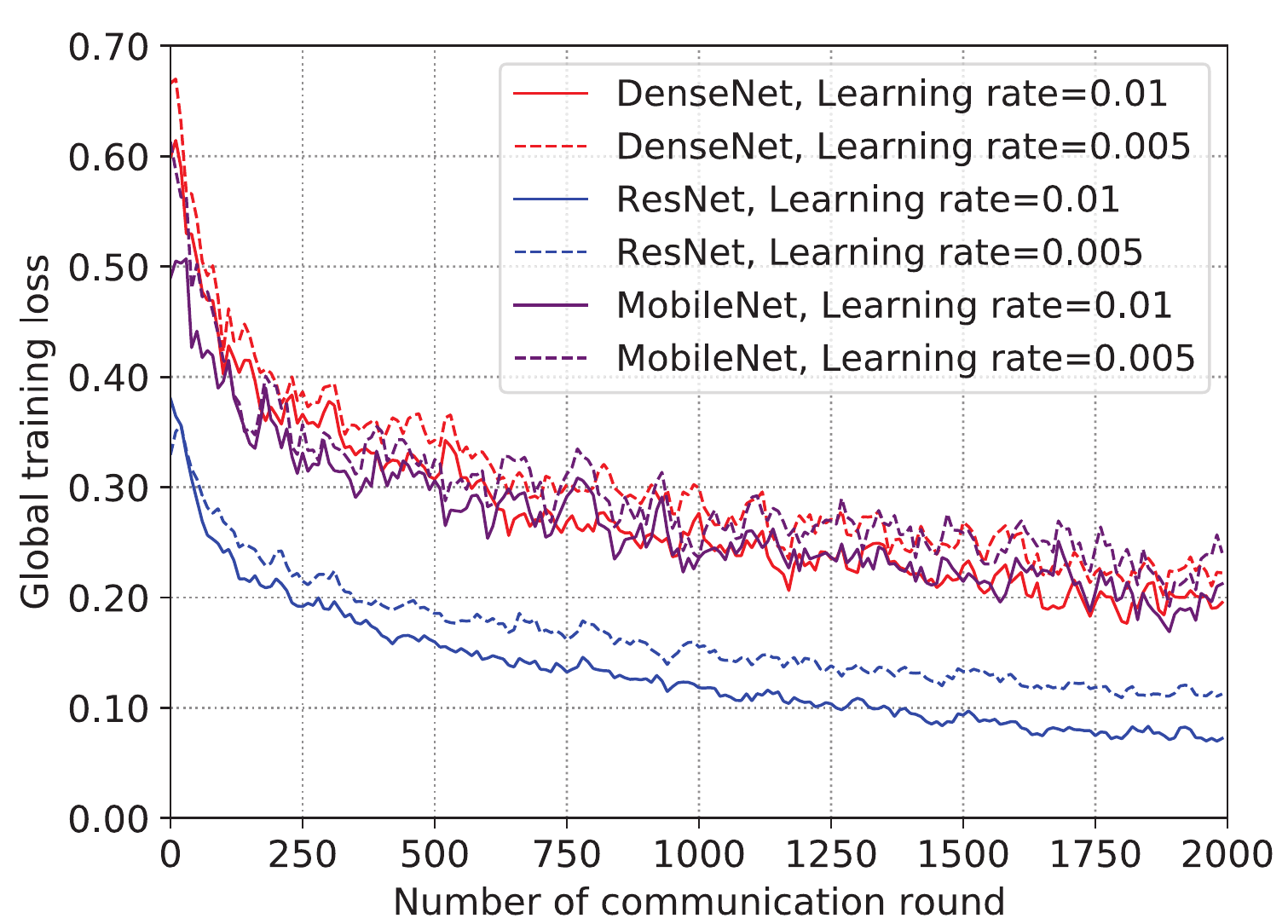}

	}
	\subfigure[Test accuracy]{
		\includegraphics[width=3in]{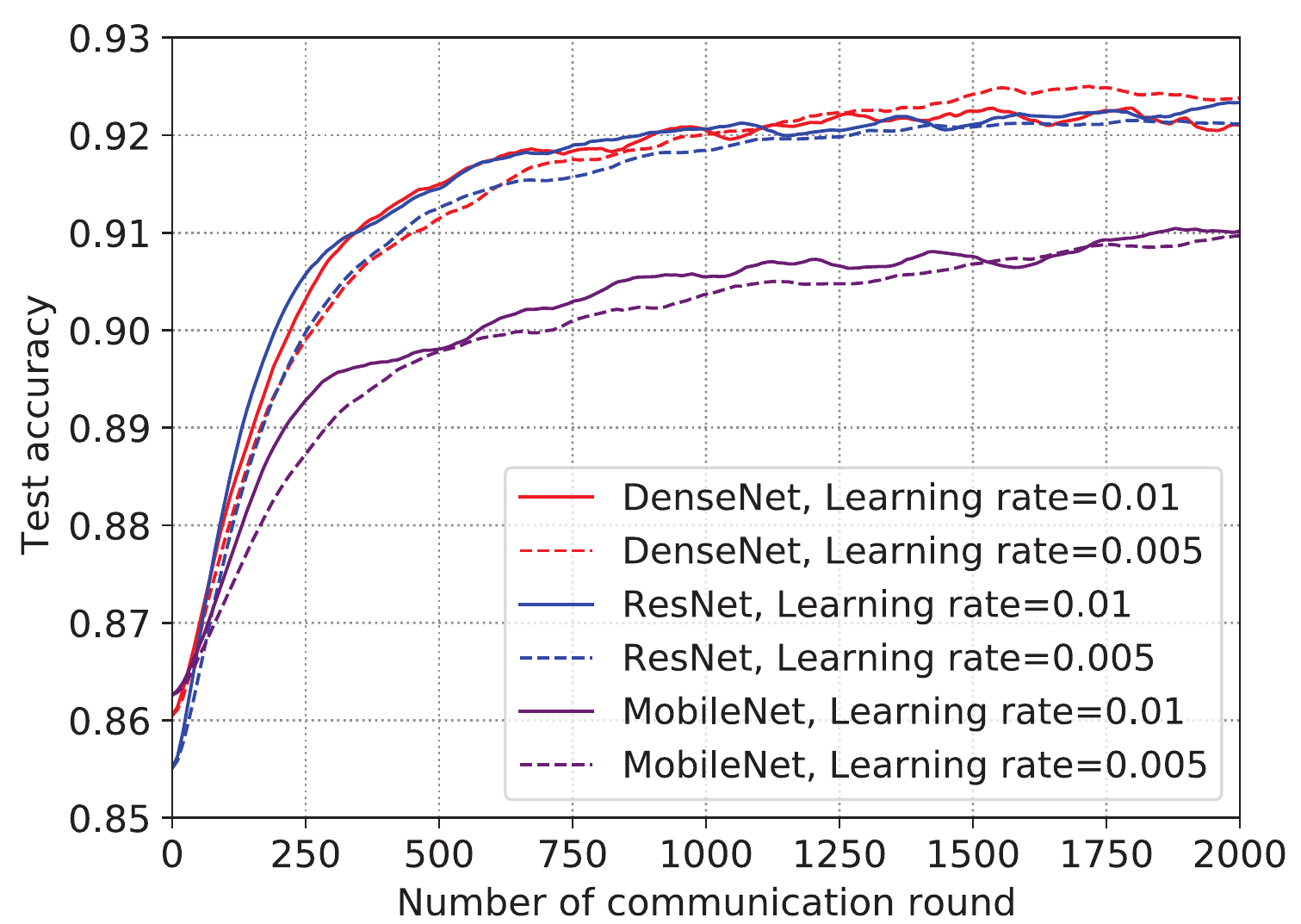}
	}
	\caption{Global training loss and test accuracy with different learning rates.}\label{Generalization_Ability}
\end{figure}
An important test of the proposed algorithm is whether it is applicable for training different DNN models, i.e., the generalization ability. Toward this end, we simulate the three DNN models in an FEEL system with $K = 12$ user devices. Specifically, the CPU frequencies of the devices are configured as: four devices with $0.7$ GHz, four devices with $1.4$ GHz, and four devices with $2.1$ GHz.

We test both the convergence speed and the learning accuracy of each DNN model with different learning rates. To be more realistic, we implement the non-IID case and the results are presented in Fig. \ref{Generalization_Ability}. It can be observed that the proposed scheme is capable of attaining a desired learning accuracy with fast convergence speed for all models. Moreover, the ultimate learning accuracy can be well guaranteed with different learning rates. This result demonstrates the strong generalization ability of the proposed algorithm, which promotes its wide implementation in practical systems.
\begin{table*}[!htp]
\caption{Training Performance of Different Schemes}
\small
\centering{}
\vspace{-0.7em}
(a) $K=6$\\
\vspace{0.5em}
\begin{tabular}{|c|c|c|c|c|}
    \hline
     \multirow{2}*{Scheme} &\multicolumn{2}{c|}{IID case} &\multicolumn{2}{c|}{Non-IID case}\\
    \cline{2-5}
    &Test accuracy &Training speedup &Test accuracy &Training speedup\\
    \cline{1-5}
    Individual learning &$90.65\%$&$1\times$&$89.16\%$&$1\times$\\
        \cline{1-5}
    Model-based FEEL &$91.28\%$&$0.29\times$&$90.22\%$&$0.37\times$\\
        \cline{1-5}
    Gradient-based FEEL &$91.60\%$&$0.53\times$&$91.51\%$&$0.69\times$\\
        \cline{1-5}
    Our proposed scheme &$91.48\%$&$1.09\times$&$91.43\%$&$1.03\times$\\
        \cline{1-5}
\end{tabular}

\vspace{1.5em}
(b) $K=12$\\
\vspace{0.5em}
\begin{tabular}{|c|c|c|c|c|}
    \hline
     \multirow{2}*{Scheme} &\multicolumn{2}{c|}{IID case} &\multicolumn{2}{c|}{Non-IID case}\\
    \cline{2-5}
    &Test accuracy &Training speedup &Test accuracy &Training speedup\\
    \cline{1-5}
    Individual learning &$90.68\%$&$1\times$&$89.91\%$&$1\times$\\
        \cline{1-5}
    Model-based FEEL &$92.01\%$&$0.32\times$&$91.16\%$&$0.31\times$\\
        \cline{1-5}
    Gradient-based FEEL &$92.27\%$&$0.68\times$&$91.81\%$&$0.67\times$\\
        \cline{1-5}
    Our proposed scheme &$92.34\%$&$1.16\times$&$92.12\%$&$1.26\times$\\
        \cline{1-5}
\end{tabular}
\end{table*}
\subsection{CPU Scenario: Performance Comparison among Different Schemes}
To demonstrate the effectiveness of the proposed scheme, we compare it with three benchmark schemes in the scenarios of $K=6$ and $K=12$ devices, respectively. Without loss of generality, we take a pre-trained DenseNet121 with $86\%$ initial learning accuracy for the following test. Both IID and non-IID cases are implemented. The detailed procedure of each scheme is summarized as follows.
\begin{itemize}
    \item \textit{Individual Learning}: Each device trains its DNN model until the local loss function converges. Then, the edge server aggregates (averages) all local models and transmits the result to the devices.
    \item \textit{Model-Based FEEL}: Each device trains its DNN model using the local dataset with one epoch. Thereafter, the model parameters are transmitted to the edge server for averaging. The result is then sent back to each device. These steps are iterated until the global model converges \cite{Communication_Efficient_from_Decentralized_Data}.
    \item \textit{Gradient-Based FEEL}: Each device trains its DNN model by running one-step SGD algorithm. The locally computed gradient is then transmitted to the edge server for averaging. After that, the global gradient is sent back to each device. These steps are iterated until the global model converges \cite{Large_Scale_DNN}.
\end{itemize}
The main difference between the proposed scheme and the gradient-based FEEL scheme is that we take into account the joint batchsize selection and communication resource allocation.

For exposition, we use two metrics to evaluate the training performance, i.e., the test accuracy and the training speedup. The training speedup is defined as the ratio of the training speed of each scheme to that of the individual learning scheme.

Table II(a) shows the training performance of all schemes in the scenario of $K = 6$ devices. As compared with the individual learning scheme, the proposed scheme can speed up the training task by about $1.09$ times with about $0.83\%$ learning accuracy improvement in the IID case, and can speed up the training task by about $1.03$ times with about $2.27\%$ learning accuracy improvement in the non-IID case.
Moreover, compared with the model-based FEEL scheme, the proposed scheme can achieve about $3.76$ times training speed in the IID case, and can achieve about $2.78$ times training speed in the non-IID case, both with a slight learning accuracy improvement. The reason is that the gradients can be deeply compressed, therefore the communication overhead for gradient transmission is much smaller than that for parameter transmission. In addition, compared with the gradient-based FEEL scheme, the proposed scheme is capable of achieving about $2$ times and $1.5$ times training speed in the IID and non-IID cases, respectively, both with almost the same learning accuracy. It is rather intuitive that the proposed scheme optimally selects the training batchsize to balance the communication and computation costs. In contrary, the gradient-based FEEL scheme trains each local model using the entire dataset without considering the communication-and-computation trade-off.

We further test the training performance of all schemes in the scenario of $K= 12$ devices. The results are illustrated in Table II(b). It can be observed that the proposed scheme can still achieve the fastest training speed as well as attain a desired learning accuracy among all schemes, which demonstrates its superiority and scalability. Moreover, as the number of devices increases, the training speed improvement of the proposed scheme become more evident than that of the individual learning scheme. It is because with more devices, the proposed scheme can exploit more computation resources to train the DNN model. In particular, the accuracy gap between IID and non-IID cases in the individual learning scheme is larger than those in other schemes. As we can imagine, the individual learning scheme cannot grasp the characteristic of the non-IID data because it performs local training without collaboration. Conversely, other three schemes periodically aggregate the local learning updates, making the accuracy gaps between IID and non-IID cases much smaller. This result also suggests the superiority of the proposed scheme to deal with the non-IID data in FEEL systems.
\begin{figure}[htp!]
	\centering
	\subfigure[Global training loss vs training time.]{
		\includegraphics[width=3in]{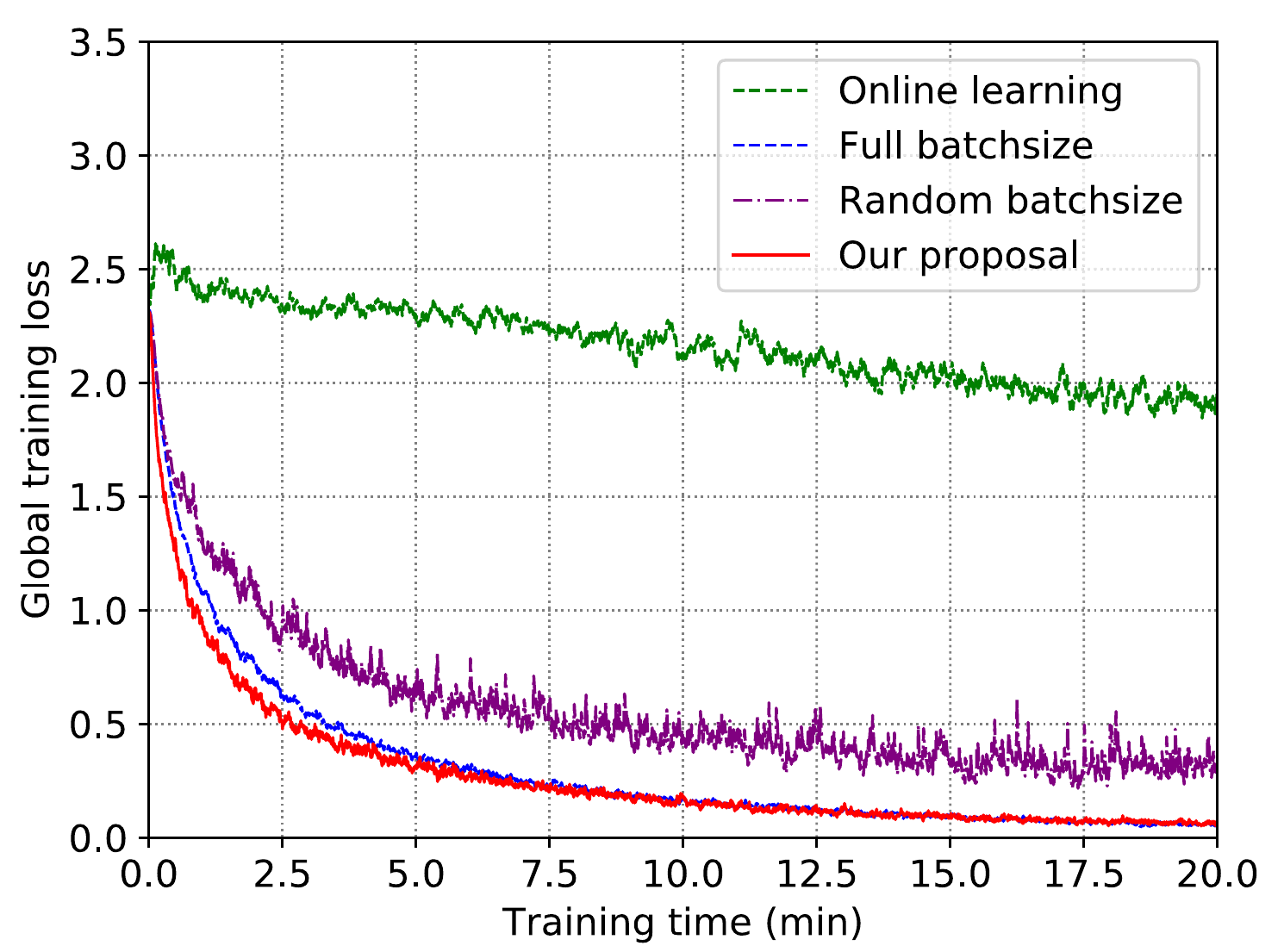}

	}
	\subfigure[Test accuracy vs training time.]{
		\includegraphics[width=3in]{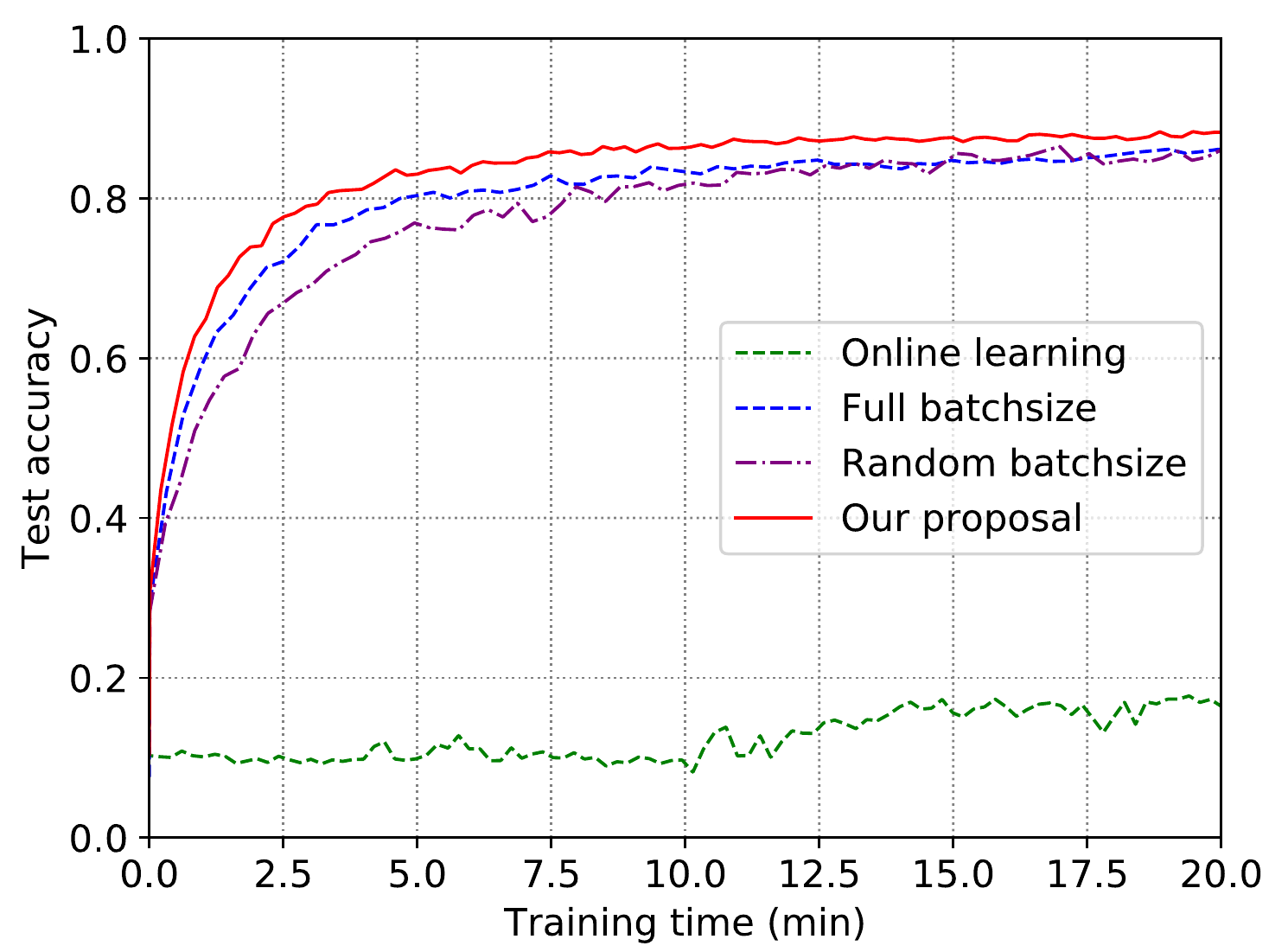}
	}
	\caption{Performance comparison among different schemes in the IID case.}\label{IID_GPU_Comparison}
\end{figure}
\begin{figure}[htp!]
	\centering
	\subfigure[Global training loss vs training time.]{
		\includegraphics[width=3in]{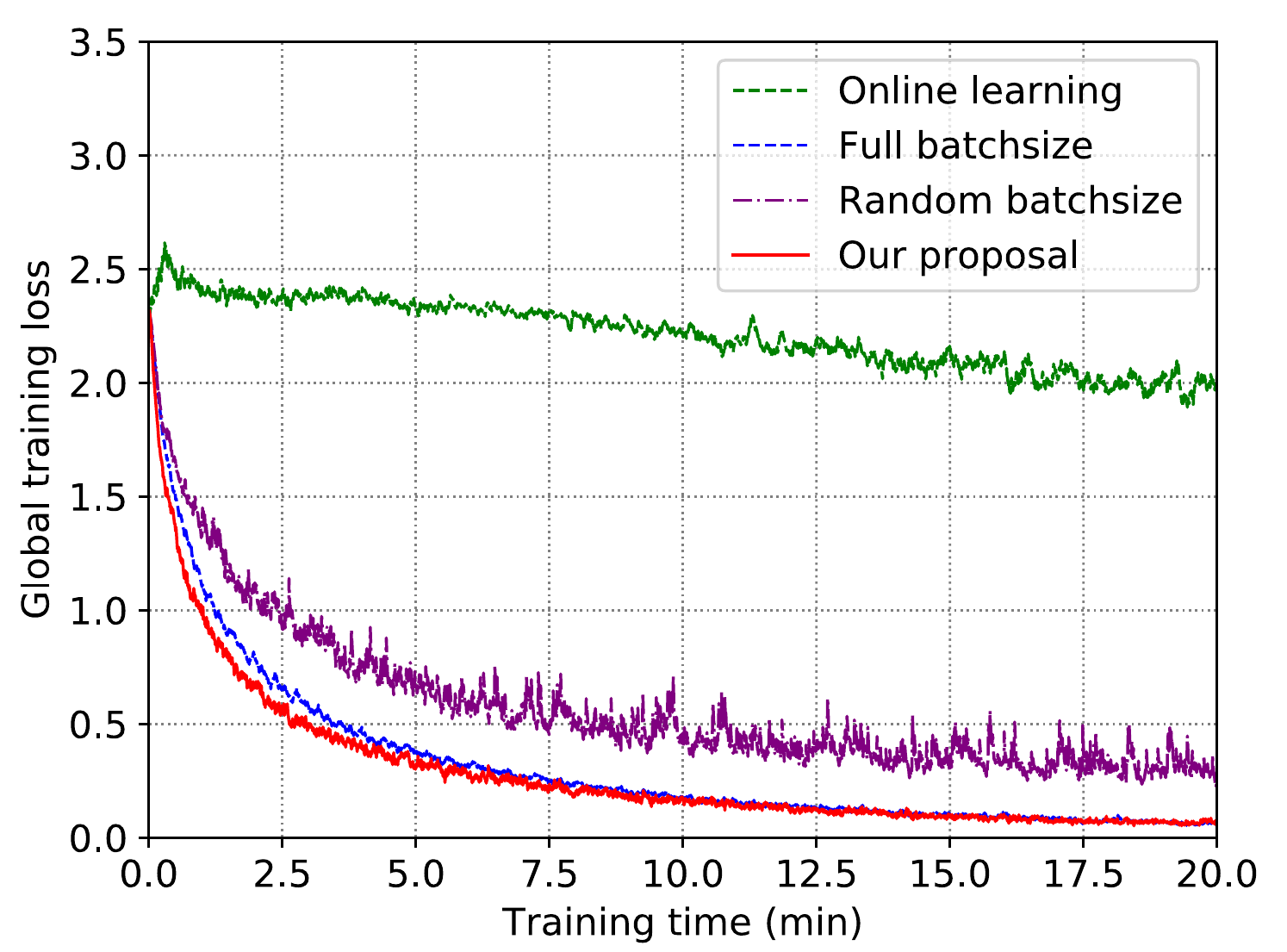}

	}
	\subfigure[Test accuracy vs training time.]{
		\includegraphics[width=3in]{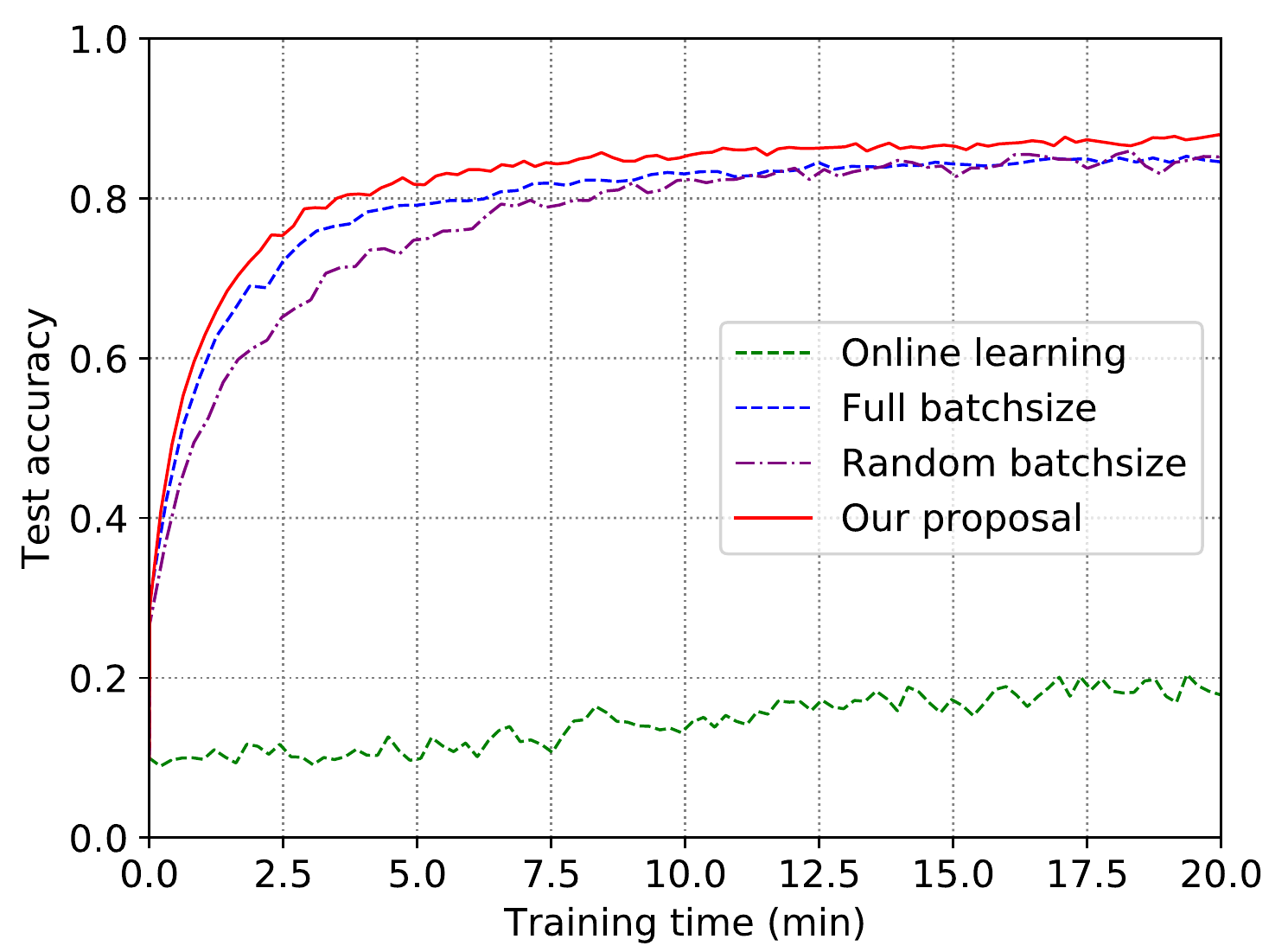}
	}
	\caption{Performance comparison among different schemes in the non-IID case.}\label{NON_IID_GPU_Comparison}
\end{figure}
\subsection{GPU Scenario: Performance Comparison among Different Schemes}
An important test is whether the proposed joint batchsize selection and communication resource allocation policy can accelerate the training task in the GPU scenario. Since the theoretical result has proved that the GPU and CPU scenarios share the similar solution, we do not replicate the preceding tests in the GPU scenario due to space limits. Alternatively, we turn to compare the training performance of the proposed scheme against three baseline schemes with different batchsizes. We consider a scenario with $K = 6$ devices, where each device is equipped with an NVIDIA GeForce GTX 1080 Ti GPU. Similarly, the DenseNet121 is tested and both IID and non-IID cases are experimented. The three baseline schemes are summarized as follows.
\begin{itemize}
    \item \textit{Online learning scheme}, where the training batchsize of each device is $B_k = 1$,~$\forall k$.
    \item \textit{Full batchsize scheme}, where the training batchsize of each device is $B_k = B^{\text{max}} = 128$,~$\forall k$.
    \item \textit{Random batchsize scheme}, where each device randomly selects a training batchsize between $1$ and $128$ in each communication round.
\end{itemize}

Fig. \ref{IID_GPU_Comparison} and Fig. \ref{NON_IID_GPU_Comparison} show the convergence speed and the learning accuracy in IID case and non-IID case, respectively. We can observe from all plots that the proposed scheme can achieve the fastest convergence speed while attaining the highest learning accuracy at the same time. The reason is that the proposed scheme can optimally allocate the time-slot resource and select appropriate training batchsize to balance the communication and computation costs. Further, the learning accuracy of the proposed scheme in the IID and non-IID cases are almost the same. It is because the edge server can aggregate the gradient information from all devices in each communication round. Therefore, the global gradient is approximately an unbiased estimate of the stochastic gradient in both IID and non-IID cases. Toward this end, the learning accuracy in the non-IID case will not deteriorate as compared with that in the IID case. This also demonstrates the applicability of the proposed scheme in practical systems with non-IID data.
\section{Conclusion}
This paper aims at accelerating the DNN training task in FEEL. The key idea is to jointly optimize local training batchsize and communication resource allocation to achieve fast training speed while maintaining learning accuracy. In the common CPU scenario, we formulate a training acceleration problem and develop a joint batchsize selection and communication resource allocation policy in closed form. Some insightful results are also discussed to provide meaningful guidelines for hypeparameter adjustment and network planning. We further extend the training acceleration problem to the GPU scenario. The optimal solution in this case can be derived by fine-tuning the result in the CPU scenario, recommending that the proposed algorithm is applicable in more general systems. Our studies in this work provide an important step towards the implementation of AI in wireless communication systems.

In this work, the training batchsize is optimized towards maximizing the learning efficiency, which may cause the global gradient being biased, especially in the non-IID data distributed system. In this regard, our analytical framework can be extended by considering the unbiased gradient as an additional batchsize constraint in $\mathscr{P}_1$. Moreover, some interesting issues, such as batchsize optimization in analogy communication systems, distributed learning with unlabeled data, and long-term training acceleration based on Markov decision process, also deserve further investigation in the future work.
\begin{appendices}
\section{Proof of Theorem 1}
According to Lemma 1, problem $\mathscr{P}_4$ is convex for a given value of $B$. Therefore, it can be solved by the Lagrange multiplier method. The  partial Lagrange function can be defined as
\begin{equation}
	\begin{aligned}
        L = E^{\text{U}} &+ \sum_{k=1}^K \lambda_k \left( \frac{B_k C}{f_k^{\text{C}}} + \frac{s T_f^{\text{U}}}{\tau_k^{\text{U}} R_k^{\text{U}}} - \xi \sqrt{B} E^{\text{U}}\right) \\&+ \mu \left( \sum_{k=1}^K \tau_k^{\text{U}} - T_f^{\text{U}} \right) + \gamma \left( \sum_{k=1}^K B_k - B \right),
    \end{aligned}
\end{equation}
where $\lambda_k$, $\mu$, and $\gamma$ are the Lagrange multipliers associated with constraints (\ref{P4b}), (\ref{P1b}), and (\ref{P1d}), respectively. Denote $\left\{B_k^*, \tau_k^{\text{U}*}\right\}$ as the optimal solution to problem $\mathscr{P}_4$. We note that the uplink communication resource $\tau_k^{\text{U}}$ is non-negative while the batchsize of each device is bounded in the interval $[1, B^{\text{max}}$]. Then, applying the KKT conditions gives the following necessary and sufficient conditions, as
\begin{align}
	&\frac{\partial L}{\partial E^{\text{U}*}} = 1 - \xi \sqrt{B}\sum_{k=1}^K \lambda_k^* = 0,\\
    &\frac{\partial L}{\partial \tau_k^{\text{U}*}} =  -\lambda_k^* \frac{ s T_f^{\text{U}}}{R_k^{\text{U}} \left(\tau_k^{\text{U}*}\right)^2} + \mu^*
        \left\{
        \begin{array}{ll}
        \geq 0,~\tau_k^{\text{U}*}=0\\
        =0,~\tau_k^{\text{U}*}>0
        \end{array},~\forall k,
	    \right.\\
    &\frac{\partial L}{\partial B_k^*} =  \lambda_k^* \frac{C}{f_k^{\text{C}}} + \gamma^*
        \left\{
        \begin{array}{ll}
        \geq 0,~B_k^*=1\\
        =0,~B_k^* \in \left( 1, B^{\text{max}}\right)\\
        \leq 0,~B_k^*=B^{\text{max}}
        \end{array}, ~\forall k,
        \right.\\
    &\lambda_k^* \left( \frac{B_k^* C}{f_k^{\text{C}}} + \frac{s T_f^{\text{U}}}{\tau_k^{\text{U}*} R_k^{\text{U}}} - \xi \sqrt{B} E^{\text{U}*}\right) = 0,~\lambda_k^* \geq 0,~\forall k,\\
    &\frac{B_k^* C}{f_k^{\text{C}}} + \frac{s T_f^{\text{U}}}{\tau_k^{\text{U}*} R_k^{\text{U}}} \leq\xi \sqrt{B}E^{\text{U}*},~\forall k,\label{inequality}\\
    &\mu^* \left( \sum_{k=1}^K \tau_k^{\text{U}*} - T_f^{\text{U}} \right) = 0,~\mu^* \geq 0,\\
    &\sum_{k=1}^K \tau_k^{\text{U}*} \leq T_f^{\text{U}}, ~~\tau_k^{\text{U}*} \geq 0,~\forall k,\\
    &\sum_{k=1}^K B_k^* = B,~~1 \leq B_k^* \leq B^{\text{max}},~\forall k.
\end{align}
With simple mathematical calculation, we can derive the optimal batchsize selection policy as
\begin{equation}
    B_k^* =  \left[ \left( \Delta L E^{\text{U}*} - \left( \frac{\Delta L s T_f^{\text{U}} \mu^* }{\rho_k R_k^{\text{U}}}\right)^{\frac{1}{2}} \right) V_k \right]_{1}^{B^{\text{max}}},~\forall k . \label{A batchsize}
\end{equation}
Moreover, $E^{\text{U}*}$ is achieved when ``$\leq$" in (\ref{inequality}) is set to ``$=$". Combining (\ref{inequality}) and (\ref{A batchsize}), we can obtain the optimal uplink time-slot allocation in Theorem 1.

\section{}
\subsection{Proof of Corollary 1}
To prove Corollary 1, we first analyze the following two cases.\\
1) \textit{Case A (Equivalent resource allocation) }

In this case, the time-slot resource allocated to each device is identical and each device selects the same training batchsize, i.e., $\tau_k^{\text{U}}=\frac{T_f^{\text{U}}}{K}$ and $B_k=\frac{B}{K},~\forall k$. As a result, the corresponding objective value in (\ref{P4a}) can be given by
\begin{align}
     E_h^{\text{U}} = \max \limits_{k \in \mathcal{K}} \left\{ \frac{1}{\Delta L} \left(\frac{B}{K V_k} + \frac{K s}{R_k^{\text{U}}} \right) \right\}. \label{A E low}
\end{align}
Since the goal of problem $\mathscr{P}_4$ is to minimize $E^{\text{U}}$, the optimal $E^{\text{U}*}$ is no greater than $E_h^{\text{U}}$, which can be regarded as an upper bound of $E^{\text{U}*}$.\\
2) \textit{Case B (Infinite memory resource)}

In this case, the memory of each device is sufficient so that the batchsize limitation (\ref{P1e}) can be relaxed. Further, the convexity of problem $\mathscr{P}_4$ can be preserved, making the KKT conditions still effective. With the similar derivations in Appendix A, we can obtain the objective value as
\begin{align}
     E_\ell^{\text{U}} = \frac{1}{\Delta L} \left( \frac{B C}{\sum_{k=1}^K f_k^{\text{C}}} + s \left(\sum \limits_{k=1}^K \sqrt{\frac{\rho_k}{R_k^{\text{U}}}}\right)^2 \right). \label{A E upp}
\end{align}
Due to that the objective value will not increase after constraint relaxation, $E^{\text{U}*}$ is no less than $E_\ell^{\text{U}}$. Thus, $E_\ell^{\text{U}}$ can be viewed as a lower bound of $E^{\text{U}*}$. Combining (\ref{A E low}) and (\ref{A E upp}), we can obtain the range of $E^{\text{U}*}$ in Corollary 1, which ends the proof.

\subsection{Proof of Corollary 2}
Similar to the proof of Corollary 1, we consider another two cases.\\
1) \textit{Case A (Online learning)}

In this case, we assume that the optimal batchsize of each device is $B_k^* = 1,~\forall k$. According to (\ref{optimal batch}), this case happens when $\left( \Delta L E^{\text{U}*} - \left( \frac{\Delta L s T_f^{\text{U}} \mu^* }{\rho_k R_k^{\text{U}}}\right)^{\frac{1}{2}} \right) V_k \leq 1,~\forall k$, resulting in
\begin{equation}
\mu^* \geq \mu_h = \max \limits_{k \in \mathcal{K}}  \left\{\dfrac{\left( \Delta L V_k E^{\text{U}*} - 1 \right)^2 \rho_k R_k^{\text{U}} }{\Delta L s T_f^{\text{U}} V_k^2}\right\}. \label{A mu upp}
\end{equation}
2) \textit{Case B (Full batch learning)}

Similar to case A, this case requires that the optimal batchsize of each device follows $B_k^* = B^{\text{max}},~\forall k$. This case happens when $\left( \Delta L E^{\text{U}*} - \left( \frac{\Delta L s T_f^{\text{U}}  \mu^* }{\rho_k R_k^{\text{U}}}\right)^{\frac{1}{2}} \right) V_k \geq B^{\text{max}},~\forall k$, leading to
\begin{equation}
\mu^* \leq \mu_\ell =  \min \limits_{k \in \mathcal{K}} \left\{\dfrac{\left( \Delta L V_k E^{\text{U}*} - B^{\text{max}}\right)^2 \rho_k R_k^{\text{U}}}{\Delta L s T_f^{\text{U}} V_k^2}\right\}. \label{A mu low}
\end{equation}

Then, when there exists one device whose batchsize is between $1$ and $B^{\text{max}}$, the value of $\mu^*$ follows $\mu_\ell \leq \mu^* \leq \mu_h$, which completes the proof.
\section{Proof of Lemma 2}
The proof of Lemma 2 is straightforward by the reduction to absurdity. It can be observed that the global loss decay $\Delta L$ is an increasing function with each training batchsize $B_k$. However, the local gradient calculation latency $t_k^{\text{L,G}}$ keeps unchanged when $1 \leq B_k \leq B_k^{\text{th}}$. Therefore, in the data bound region, the global loss decay (numerator of (\ref{P6a})) increases with $B_k$ while the end-to-end latency (denominator of (\ref{P6a})) remains unchanged, resulting in an increasing learning efficiency. Since the objective of problem $\mathscr{P}_6$ is to maximize the learning efficiency, the optimal batchsize will not locate in the data bound region. This ends the proof.
\end{appendices}

\begin{IEEEbiography}[{\includegraphics[width=1in,clip]{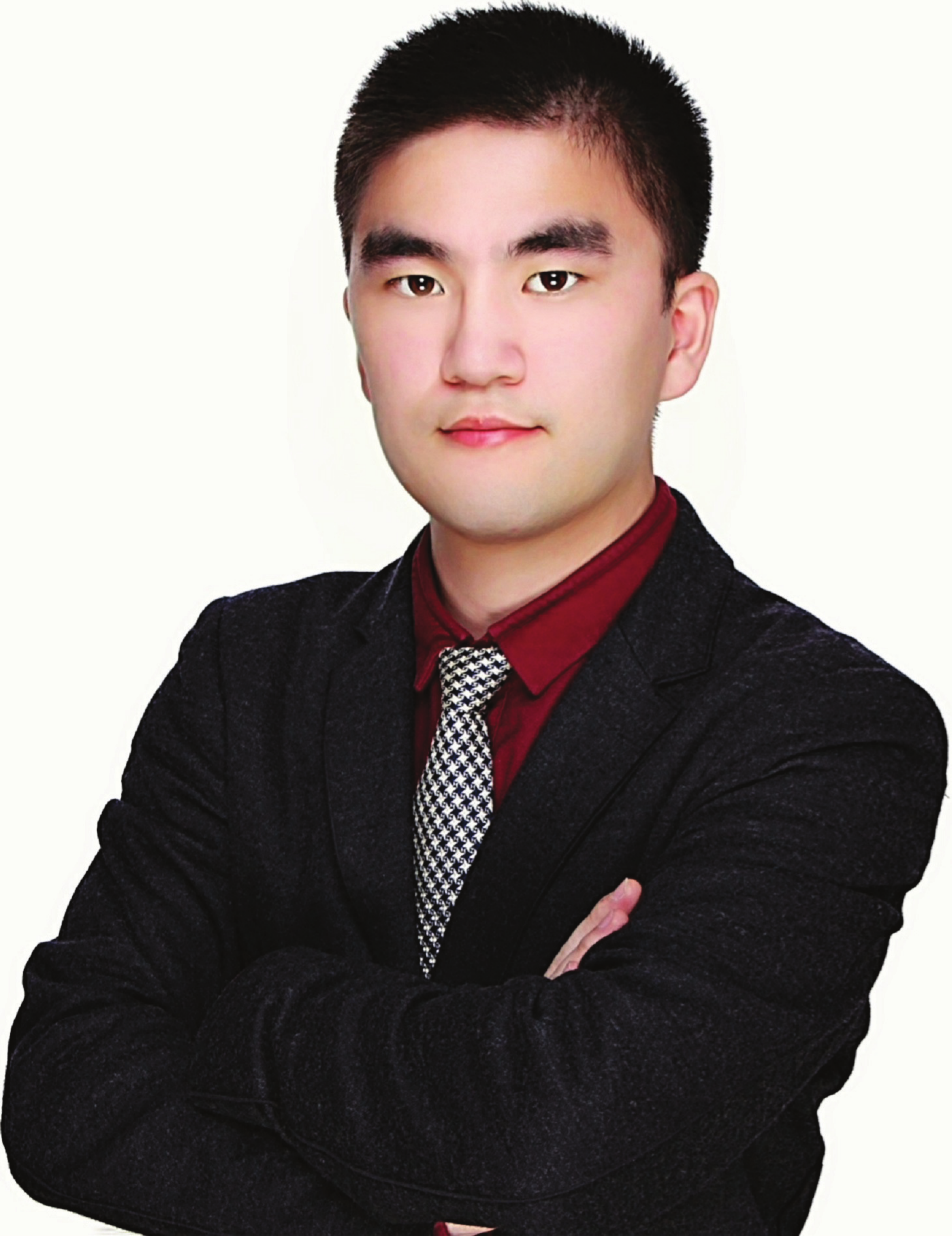}}]{Jinke Ren}(Graduate Student Member, IEEE) received the B.S.E degree in information engineering from Zhejiang University, Hangzhou, China, in 2017, where he is currently pursuing the Ph.D. degree with the College of Information Science and Electronic Engineering. Since December 2019, he has been working as a Visiting Research Scholar with the Department of Electrical and Computer Engineering, Northwestern University, IL, USA. His research interests include wireless edge intelligence, distributed machine learning, and 5G technologies such as mobile edge computing and device-to-device communication.
\end{IEEEbiography}

\begin{IEEEbiography}[{\includegraphics[width=1in,clip]{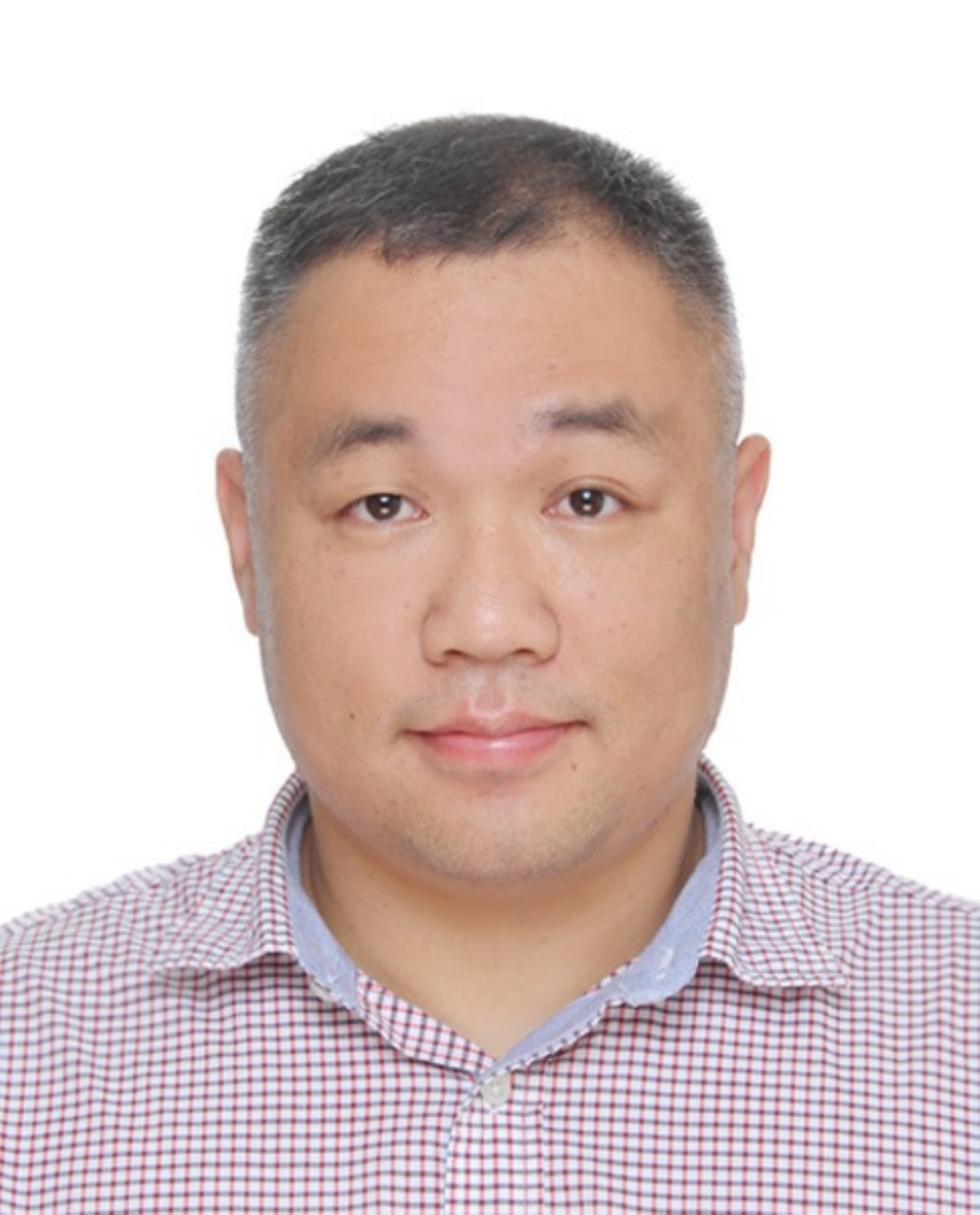}}]{Guanding Yu} (Senior Member, IEEE) received the B.E. and Ph.D. degrees in communication engineering from Zhejiang University, Hangzhou, China, in 2001 and 2006, respectively. He joined Zhejiang University in 2006, and is now a Full Professor with the College of Information and Electronic Engineering. From 2013 to 2015, he was also a Visiting Professor at the School of Electrical and Computer Engineering, Georgia Institute of Technology, Atlanta, GA, USA. His research interests include 5G communications and networks, mobile edge computing, and machine learning for wireless networks.
	
Dr. Yu has served as a guest editor of \textsc{IEEE Communications Magazine} special issue on Full-Duplex Communications, an Editor of \textsc{IEEE Journal on Selected Areas in Communications} Series on Green Communications and Networking, and a lead Guest Editor of \textsc{IEEE Wireless Communications Magazine} special issue on LTE in Unlicensed Spectrum, and an Editor of IEEE Access. He is now serving as an Editor of \textsc{IEEE Transactions on Green Communications and Networking}, an Associate Editor of \textsc{IEEE Journal on Selected Areas in Communications} Series on Machine Learning in Communications and Networks, and an Editor of \textsc{IEEE Wireless Communications Letters}. He received the 2016 IEEE ComSoc Asia-Pacific Outstanding Young Researcher Award. He regularly sits on the technical program committee (TPC) boards of prominent IEEE conferences such as ICC, GLOBECOM, and VTC. He has served as a Symposium Co-Chair for IEEE Globecom 2019 and a Track Chair for IEEE VTC 2019'Fall.
\end{IEEEbiography}

\begin{IEEEbiography}[{\includegraphics[width=1in,clip]{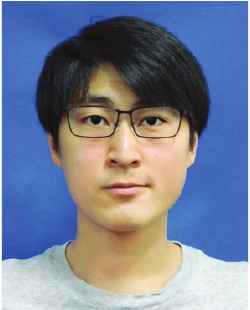}}]{Guangyao Ding} received the B.S.E degree in information engineering from Zhejiang University, Hangzhou, China, in 2019, where he is currently pursuing the Ph.D degree with the College of Information Science and Electronic Engineering. His research interests include URLLC and machine learning for wireless networks.
\end{IEEEbiography}

\end{document}